\documentclass[fleqn, 3p]{elsarticle}

\usepackage{subcaption}
\usepackage{graphicx}
\usepackage{svg}
\usepackage{amssymb,amsmath,stmaryrd,amsthm}
\usepackage{mathtools}
\usepackage{hyperref}
\usepackage[utf8]{inputenc}
\usepackage{xcolor}
\usepackage{bbold}
\usepackage{upgreek}
\usepackage{mathrsfs}
\usepackage{cleveref}
\usepackage{upgreek}
\usepackage{mathabx}
\usepackage{algorithm}
\usepackage{algpseudocode}

\biboptions{numbers,sort&compress}

\newtheorem{remark}{Remark}

\algrenewcommand\algorithmicrequire{\textbf{Input:}}
\algrenewcommand\algorithmicensure{\textbf{Output:}}

\hypersetup{
    colorlinks=true,
    linkcolor=blue,
    filecolor=magenta,      
    urlcolor=cyan,
}

\newcommand{\bfc}{\boldsymbol{c}}

\newcommand{\bff}{\boldsymbol{f}}
\newcommand{\bfg}{\boldsymbol{g}}

\newcommand{\bfs}{\boldsymbol{s}}

\newcommand{\bfu}{\boldsymbol{u}}

\newcommand{\bfx}{\boldsymbol{x}}

\newcommand{\bfV}{\boldsymbol{V}}

\newcommand{\bfX}{\boldsymbol{X}}

\newcommand{\bs}[1]{\boldsymbol{#1}}

\journal{}
 \makeatletter
 \def\ps@pprintTitle{%
   \let\@oddhead\@empty
   \let\@evenhead\@empty
   \def\@oddfoot{\reset@font\hfil\thepage\hfil}
   \let\@evenfoot\@oddfoot
 }
 \makeatother

\begin{document}

\begin{frontmatter}

\title{Learning Latent Space Dynamics with Model-Form Uncertainties: A Stochastic Reduced-Order Modeling Approach}

\author[addressMEMS]{Jin Yi Yong}
\author[addressUTA]{Rudy Geelen}
\author[addressMEMS]{Johann Guilleminot}

\address[addressMEMS]{Department of Mechanical Engineering and Materials Science, Duke University, NC, United States}
\address[addressUTA]{Oden Institute for Computational Engineering and Sciences, University of Texas at Austin, TX, United States}

\begin{abstract}
This paper presents a probabilistic approach to represent and quantify model-form uncertainties in the reduced-order modeling of complex systems using operator inference techniques. Such uncertainties can arise in the selection of an appropriate state-space representation, in the projection step that underlies many reduced-order modeling methods, or as a byproduct of considerations made during training, to name a few. Following previous works in the literature, the proposed method captures these uncertainties by expanding the approximation space through the randomization of the projection matrix. This is achieved by combining Riemannian projection and retraction operators---acting on a subset of the Stiefel manifold---with an information-theoretic formulation. The efficacy of the approach is assessed on canonical problems in fluid mechanics by identifying and quantifying the impact of model-form uncertainties on the inferred operators. 
\end{abstract}

\end{frontmatter}

\section{Introduction}
\label{sec:introduction}

Constructing efficient and robust surrogate models for decision-making and engineering system design is a critical task in computational science and engineering. In this paper, we outline a generally applicable approach for representing and propagating \emph{model-form} uncertainties by leveraging principles and tools from reduced-order modeling, Riemannian optimization and information theory. This type of uncertainty arises from the inherent limitations of approximating complex physical phenomena using mathematical models. They originate from simplifying assumptions \cite{Heyse2021}, lack of knowledge \cite{heSurveyUncertaintyQuantification2023, Farhat2018, Grandhi2014}, and/or specification of a training strategy \cite{osti_1733262, gawlikowskiSurveyUncertaintyDeep2023a}, among others. By construction, such uncertainties cannot be captured in a parametric setting \cite{bookUQSoize} and require ad hoc stochastic representations that encode the action of the underlying learned model. The methodology presented in this paper extracts, from data, knowledge and intuition that is not captured by a conventional, deterministic computational model and incorporates it into the modeling approach in a non-intrusive fashion.

Projection-based model order reduction (PMOR) typically leads to models that preserve essential dynamics yet provide significant computational savings \cite{doi:10.1137/130932715}. Its underlying techniques, along with many of their machine learning counterparts, have gained significant traction across various fields, for example---in solid mechanics \cite{Zahr2017, He2020, Guo2024}, structural dynamics \cite{Farhat2014, He2020} and fluid dynamics \cite{Grimberg2020, Grimberg2021,Tezaur2022,Stabile2018}. Despite their widespread application, these methods are often \textit{intrusive} in that they require access to the governing equations in high-fidelity models, which can be prohibitive in practice. This is especially problematic when dealing with complex simulation models and legacy codes. In contrast, \textit{non-intrusive} data-driven techniques have provided a powerful alternative for system identification and model learning tasks. Operator inference (OpInf), introduced by Peherstorfer and Willcox \cite{peherstorferDatadrivenOperatorInference2016}, is one such method that has gained attention for its ability to infer ROMs through a set of learned low-dimensional matrix operators without requiring access to the model that generated the training data. This is particularly useful when evaluating the high-fidelity model is computationally demanding or its source code is simply unavailable \cite{Ghattas2021}. In the OpInf framework the matrix operators are learned from snapshot data using least-squares regression. This makes the approach both flexible and scalable for a broad range of reduced-order modeling applications.

The original formulation of the OpInf methodology calls for a proper orthogonal decomposition (POD) to be applied to the data matrix to yield an orthonormal basis that spans a subspace of fixed dimensionality \cite{peherstorferDatadrivenOperatorInference2016}. The POD provides an optimal \emph{linear} embedding of our original high-dimensional data set into a reduced-dimensional linear subspace \cite{sirovich1987turbulence, doi:10.1137/130932715}. These methods thus face significant challenges when applied to problems in which the Kolmogorov N-width decreases slowly with increasing subspace dimension \cite{Fick2018}. The Kolmogorov-width is a mathematical measure characterizing the system's reducibility through its best linear approximation error. Limitations surrounding the use of linear subspaces become particularly problematic in systems with strong nonlinearities, moving discontinuities, or parametric dependencies. To address this issue, nonlinear MOR methods that utilize machine learning methods have been successfully applied to several applications in the field of reduced-order modeling. Despite the success of neural-network-based nonlinear MOR techniques, these methods often require significant computational resources and sizable amount of data for training \cite{leeModelReductionDynamical2020, romorNonlinearManifoldReducedOrder2023, chenCROMContinuousReducedOrder2023, frescaPODDLROMEnhancingDeep2022a}. In contrast, various recent works \cite{BARNETT2022111348, GEELEN2023115717, 10384209, geelenLearningPhysicsbasedReducedorder2023} have studied a nonlinear manifold-based approach that enriches state approximations with low-order polynomial terms, leading to improved accuracy and stability of ROMs. We will follow the principles behind the latter in the development of our probabilistic approach.

\begin{figure}[tbp]
    \centering
    \includegraphics[trim = {0 2.5cm 0 0.5cm}, clip, width = 0.8\textwidth]{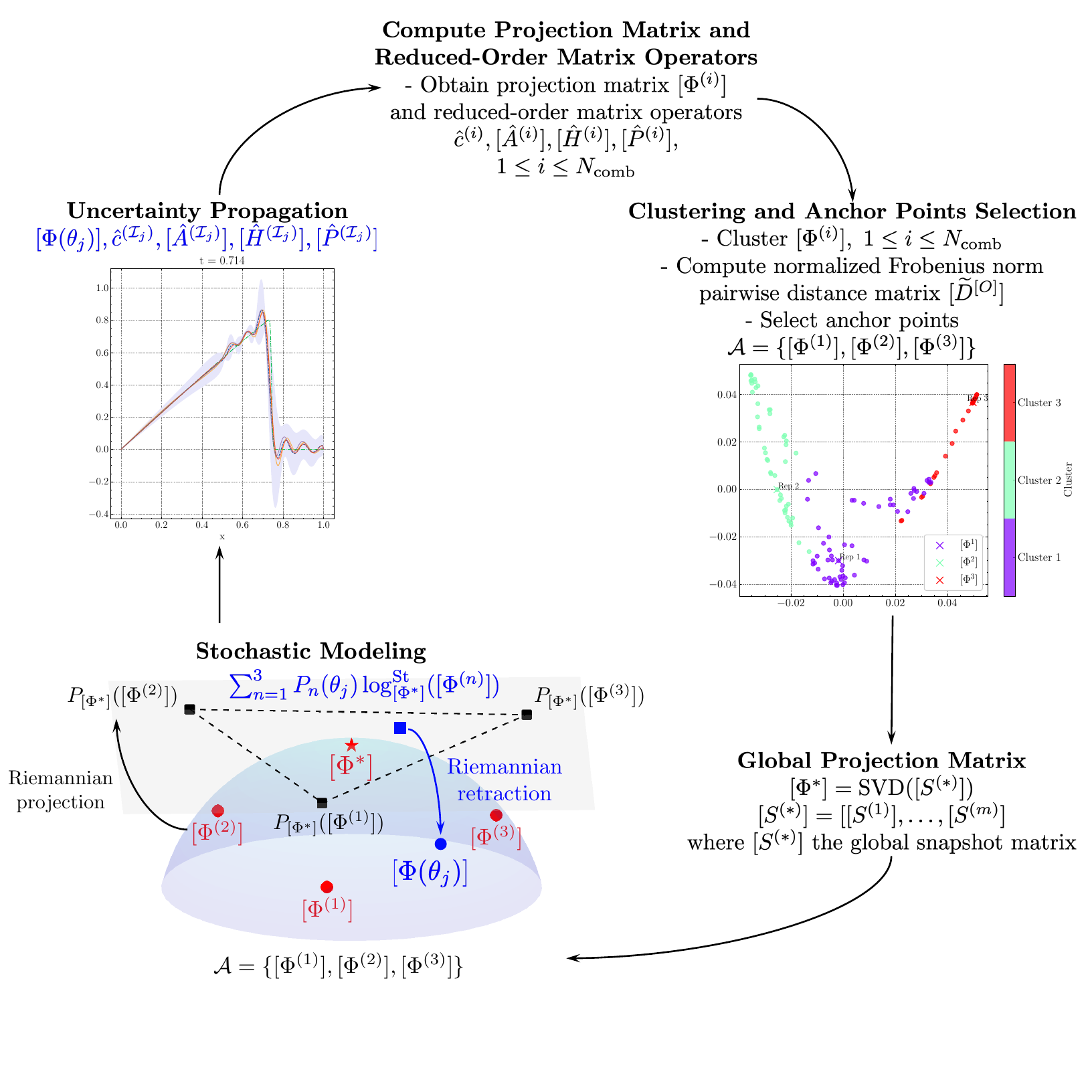}
    \caption{Overview of the stochastic reduced-order modeling approach operating in the Operator Inference (OpInf) framework. Model-form uncertainties are generated using a randomization of the projection matrix, and parameters in the reduced-order approximation are selected on the fly to propagate uncertainties to quantities of interest.}
    \label{fig: proposedApproach}
\end{figure}

Several contributions aiming at establishing probabilistic versions of data-driven ROMs (to accommodate noisy data, for instance) have been proposed recently. Perhaps the most natural step towards the integration of uncertainties consists in casting the calibration task as a Bayesian inverse problem; see \cite{GUO2022115336} for example, as well as \cite{mcquarrie2024bayesianlearninggaussianprocesses} for a Bayesian formulation leveraging enrichment through Gaussian processes. The work by Uy et al.\ \cite{uyActiveOperatorInference2023} has addressed operator inference with noisy data. Progress has also been made with regard to probabilistic error estimates; see \cite{uyProbabilisticErrorEstimation2021} for linear OpInf ROMs in the case reduced-order operators are equivalent to those obtained from explicit projection-based methods, as in \cite{peherstorferSamplingLowdimensionalMarkovian2020}. Note that providing an exhaustive review on UQ methods in the more general context of scientific machine learning (SciML), or for other classes of operator approximations (using, e.g., deep learning), is beyond the scope of this work; see \cite{PSAROS2023111902} for a general review articulating SciML and UQ, \cite{Yang2022,Garg2023,Zou2024} for deep-learning formulations embedding UQ aspects, and \cite{Fan2023} for Bayesian nonlocal operator regression, to list a few. 

This work considers the nonlinear manifold-based Operator Inference (OpInf) framework proposed in \cite{geelenLearningPhysicsbasedReducedorder2023} and seeks to enhance its robustness by allowing for the probabilistic description of uncertainties in predictions as the training strategy is being refined, e.g., as the parametric combinations indexing the training dataset vary in both values and cardinality. Since each combination defines a given reduced-order operator, we interpret the above uncertainties as model-form uncertainties and thus aim to construct a probabilistic model in a nonparametric setting. Building upon recent developments pertaining to stochastic reduced-order modeling for nonlinear dynamical systems \cite{soize2017probabilistic,soize2019probabilistic,zhangRiemannianStochasticRepresentation2023a} and coupled partial differential equations
\cite{zhangRepresentingModelUncertainties2024}, we specifically develop a methodological extension to the OpInf formulation that combines the randomization of the projection matrix, which effectively expands the approximation space, with a probabilistic model defined in the context of information theory \cite{Jaynes1957a,Jaynes1957b}. In what follows, this model will be referred to as the information-theoretic model. The proposed method enables the sampling of model perturbations with proper concentration in the convex hull defined by a set of model proposals, identified by variations in the selection of training parameters for instance. This strategy offers several key advantages:
\begin{enumerate}
\item It provides a natural framework for capturing model-form uncertainties through the projection matrix, which directly impacts the action of the learned operator.
\item The use of Riemannian operators ensures that the randomized matrices respect the underlying manifold structure, maintaining both physical consistency and dynamical stability (in the almost sure sense).
\item The information-theoretic formulation allows for a principled approach to probabilistic representation, minimizing bias by leveraging the principle of maximum entropy.
\item The resulting stochastic framework is computationally efficient and can easily be integrated within existing OpInf workflows.
\end{enumerate} 
By addressing the representation of model-form uncertainties in OpInf methods, this work contributes to the broader goal of enhancing the robustness of learning approaches for reduced-order models in complex physics-based systems. In doing so, it complements other contributions focusing on model error correction (see, e.g., \cite{CAO2023112104,JHA2024116595}) and optimal architecture selection (see \cite{SINGH2024117061} for a contribution in the context of Bayesian model selection, for instance), developed for deep learning models. An overview of the proposed probabilistic framework is shown in Fig.~\ref{fig: proposedApproach}.

The remainder of this paper is organized as follows. Section \ref{subsec:PrototypicalForm} reviews the OpInf framework for constructing nonlinear state-space representations using manifolds and representing the dynamics of the system of interest on these manifolds. We then outline the probabilistic formulation for model-form uncertainties, using probability theory, in Section \ref{subsec:SROM}. This includes a brief discussion on the use of Riemannian optimization for randomizing the projection matrix. In Section \ref{sec:numerical_results} our methodology is applied to the Burgers' equation and the incompressible Navier-Stokes equations, showcasing the parametric analysis of the deterministic ROM and uncertainty quantification results using our stochastic framework. Section \ref{sec:conclusions} provides concluding remarks, discusses limitations, and outlines potential future research directions.\\

\noindent \textbf{Notation}\\
A lower-case Latin or Greek letter, such as $x$ or $\eta$, is a deterministic real variable.\\
A boldface lower-case Latin or Greek letter, such as $\bfx$ or $\bs{\eta}$, is a deterministic vector.\\
An upper-case Latin or Greek letter, such as $X$ or $\Xi$, is a real-valued random variable.\\
A boldface upper-case Latin letter, such as $\bfX$, is a vector-valued random variable.\\
A lower- or upper-case Latin letter between brackets, such as $[x]$ or $[X]$, is a deterministic matrix.\\
A boldface upper-case letter between brackets, such as $[\bfX]$, is a matrix-valued random variable.

\section{Formulation for Learning Latent Space Dynamics on Nonlinear Manifolds}
\label{subsec:PrototypicalForm}

This section first presents a general methodology for low-dimensional approximation using nonlinear manifolds. In particular, Section~\ref{subsec:nonlinear_representations} reviews the methodology from \cite{geelenLearningPhysicsbasedReducedorder2023} for learning nonlinear representations through the use of polynomial feature maps. We then discuss the computation of reduced-order models in which the latent space dynamics can be expressed \emph{and} learned in terms of this new representation in Section~\ref{subsec:latent_space_dynamics}.

\subsection{Learning Nonlinear Representations}
\label{subsec:nonlinear_representations}
Consider a training dataset comprised of a set of snapshots corresponding to the high-dimensional state-space variable (e.g.~temperature or pressure) in a physical system of interest. We denote each snapshot by $\bfs_j \in \mathbb{R}^{N}$, and assemble a snapshot matrix $[S] \in \mathbb{R}^{N \times k}$ from $k$ such snapshots:
\begin{equation}
    [S] = \left(\begin{array}{cccc}
        \mid & \mid & & \mid \\
        \bfs_1 & \bfs_2 & \ldots & \bfs_k \\
        \mid & \mid & & \mid
        \end{array}\right)\,.
\end{equation}
Note that $[S]$ may contain states from multiple full model evaluations corresponding to different initial conditions and/or boundary conditions. To enable the derivation of the probabilistic formulation in a rather general setting, we consider a low-dimensional approximation $\Gamma: \mathbb{R}^r \mapsto \mathbb{R}^N$ to ${\bfs}(t) \in \mathbb{R}^N$ defined as 
\begin{equation}
    {\bfs}(t) \approx \Gamma(\hat{\bfs}(t)) = {\bfs}_{\text{ref}} + [V_r]\hat{\bfs}(t) + [\overline{V}]\mathcal{N}(\hat{\bfs}(t))\,,
    \label{eqn: nonlinear_rep}
\end{equation}
where $\hat{\bfs}(t) \in \mathbb{R}^r$ is the reduced-order state vector at time $t$, defined as the coefficients of the reduced model in the low-dimensional (linear subspace) basis,  ${\bfs}_{\text{ref}}$ is a reference state vector (defined through, e.g., an initial condition or the mean system-state), $[V_r] \in \mathbb{R}^{N \times r}$ and $[\overline{V}] \in \mathbb{R}^{N \times q}$ are orthonormal matrices such that $[[V_r],[\overline{V}]] \in \mathbb{R}^{N \times (r+q)}$ is a projection matrix computed from the snapshot matrix $[S]$, with $(r+q) \ll N$ typically, and $\mathcal{N}: \mathbb{R}^r \mapsto \mathbb{R}^{q}$ defines a nonlinear enrichment for the state-space approximation. 

Several choices for the enrichment function $\mathcal{N}$ have been proposed in the literature (see Remark~\ref{rem: remarkNNIntrusive} at the end of this section). Here, we consider the polynomial approximation defined in \cite{geelenLearningPhysicsbasedReducedorder2023} as
\begin{equation}
    \mathcal{N}(\hat{s}(t)) = [\Xi] \bfg(\hat{\bfs}(t))\,,
\end{equation}
where the coefficient matrix $[\Xi] \in \mathbb{R}^{q \times (p-1)r}$ weights the contributions of the singular vectors in $[\overline{V}]$ and the vector $\bfg(\hat{\bfs}(t)) \in \mathbb{R}^{(p-1)r}$ has components $\hat{\bfs}^j(t) = [\hat{\bfs}_1^j(t), \hat{\bfs}_2^j(t), \dots, \hat{\bfs}_r^j(t)]^\top$, with $j \in [2, \dots, p]$. The snapshots of the reduced-order state-space variable $\hat{\bfs}(t)$ are obtained by solving the following constrained optimization problem 
\begin{equation}
        \underset{[V], [\overline{V}], {[\Xi]}, \hat{[S]}}{\arg \min}
        \quad F\left([V], [\overline{V}], [\Xi], [\hat{S}]\right)+\frac{\gamma}{2}\|{[\Xi]}\|_F^2\,, \text{ subject to } \left([V],[\overline{V}]\right) \in \text{St}(N,(r+q))\,,
    \label{eqn: representationProb}
\end{equation}
where $[\hat{S}] = (\hat{\bfs}_1, \hat{\bfs}_2, \dots, \hat{\bfs}_k) \in \mathbb{R}^{r \times k}$ collects the snapshots of the reduced-order state-space representation, $F$ is the objective function defined as 
\begin{equation}
\begin{aligned}
        F([V], [\overline{V}], [\Xi], [\hat{S}]) &= \frac{1}{2}\sum_{j=1}^k ||\bfs_j - \Gamma(\hat{\bfs}_j) ||_2^2 \\ &= \frac{1}{2}\sum_{j=1}^k \left\| \bfs_j - {\bfs}_{\text{ref}} - \left([V], [\overline{V}]\right) \left( \begin{array}{c}
                \hat{\bfs}_j \\
                \mathcal{N}(\hat{\bfs}_j)
        \end{array} \right) \right\|_2^2\,,        
    \label{eqn: representationObj}
\end{aligned}
\end{equation}
$\gamma$ is a regularization parameter, and $\text{St}(N, (r+q))$ is the Stiefel manifold (which is the set of $N$-by-$(r+q)$ orthonormal matrices; see the reference paper \cite{Edelman1998}, as well as \cite{Absil-2007}, for details). The $\gamma$-regularization term in Eq.~\eqref{eqn: representationProb} is introduced to avoid overfitting \cite{geelenLearningPhysicsbasedReducedorder2023}. The representation learning problem \eqref{eqn: representationProb}--\eqref{eqn: representationObj} can be solved using a number of techniques (see \cite{10384209, geelenLearningPhysicsbasedReducedorder2023}, for instance). 
 
It should be noted that linear approximations given by projections onto the leading $r$ principal components may lead to poor results in combination with correction terms because the linear subspace is \emph{agnostic} to the nonlinear feature map. This problem may be alleviated by selecting principal components that are not necessarily ordered in descending order with respect to the singular values. The greedy method introduced in \cite{schwerdtner2024greedyconstructionquadraticmanifolds} demonstrates that such a strategy has the potential to increase the representation accuracy by several orders of magnitude without compromising on scalability or computational efficiency. This work, however, employs the first $r$ leading principal components of the training data for building the linear part of the approximation. This is done to isolate and analyze the effects of uncertainties in the reduced basis construction.

\subsection{Learning Latent Space Dynamics}
\label{subsec:latent_space_dynamics}
We now target complex physics-based models governed by systems of nonlinear partial differential equations. After spatial discretisation, the dynamical-system model for the problem of interest can be expressed in semi-discrete form as
\begin{equation}
    \frac{d{\bfs}}{dt} = \mathbf{f}(\bfs)\,, \quad \bfs(0) = \bfs_0\,,
\end{equation}
where $\mathbf{f}: \mathbb{R}^N \rightarrow \mathbb{R}^N$ maps the system state $\bfs$ (with values in $\mathbb{R}^{N}$) to its time derivative and $\bfs_0 \in \mathbb{R}^N$ is a given initial condition. In the general case where $\mathbf{f}$ is non-polynomial, variable transformations may be introduced that promote a system structure that is more amenable to projection-based approximation. This procedure is often referred to as \emph{lifting} and was described in \cite{doi:10.1137/14097255X, Kramer2019, Qian2020, Ghattas2021}. It has been demonstrated in \cite{Gu2011} that many engineering systems are characterized by nonlinear terms that may be lifted to a linear-quadratic form:
\begin{equation}
    \frac{d{\bfs}}{dt} = [A]\bfs + [H](\bfs \otimes \bfs)\,;\quad \bfs(0) = \bfs_0\,,
\end{equation}
where $\otimes$ denotes the Kronecker product\footnote{The Kronecker product is defined as an operation on two matrices of arbitrary size. In our case, the matrix $\bfs$ consists of a single column representing the system state. Acknowledging the slight abuse of notation, the Kronecker product $\bfs \otimes \bfs$ can be viewed as vectorization of the outer product.}, $[A] \in \mathbb{R}^{N\times N}$, and $[H] \in \mathbb{R}^{N\times N^2}$. By incorporating the nonlinear state approximation \eqref{eqn: nonlinear_rep} into the above equation, and applying a Galerkin projection, an approximation of the latent dynamics with polynomial structure is obtained:
\begin{equation}
    \frac{d \hat{\bfs}}{dt} = \hat{\bfc} + [\hat{A}] \hat{\bfs} + [\hat{H}](\hat{\bfs} \otimes \hat{\bfs}) + [\hat{P}] \hat{\bfg}(\hat{\bfs}), \quad \hat{\bfs}(0) = \hat{\bfs}_0\,,
\end{equation}
where $\hat{\bfc} \in \mathbb{R}^r$, $[\hat{A}] \in \mathbb{R}^{r \times r}$, $[\hat{H}] \in \mathbb{R}^{r \times r^2}$, and $[\hat{P}] \in \mathbb{R}^{r \times d(r,p)}$ define reduced-order matrices, and $\hat{\bfs}(0) = \hat{\bfs}_0$ is the representation of initial condition $\bfs_0$ in the latent space. The vector $\hat{\bfg}(\hat{\bfs})$ is a subvector of $\hat{\bfs}$ (of length $d(r,p)$) and contains monomials of degree three up to degree $2p$. The matrix $[\hat{P}]$ accounts for higher-order interactions between the reduced-order state coefficients, captured by the vector $\hat{\bfg}(\hat{\bfs})$. Details regarding its construction can be found in \cite{geelenLearningPhysicsbasedReducedorder2023}.
 
The vector $\hat{\bfc}$ and matrices $[\hat{A}]$, $[\hat{H}]$ and $[\hat{P}]$ defining the reduced-order operator are obtained as the solution to the following regularized least-squares problem:
\begin{equation}
        (\hat{\bfc}, [\hat{A}], [\hat{H}], [\hat{P}]) = \underset{\tilde{\bfc}, [\tilde{A}], [\tilde{H}], [\tilde{P}]}{\arg \min } \quad J(\tilde{\bfc}, [\tilde{A}], [\tilde{H}], [\tilde{P}]) +\frac{\lambda_1}{2}\left(\|\tilde{\bfc}\|_2^2+\|[\tilde{A}]\|_F^2\right)+\frac{\lambda_2}{2}\|[\tilde{H}]\|_F^2+\frac{\lambda_3}{2}\|[\tilde{P}]\|_F^2\,,
    \label{eqn: OpInfLeastSquares}
\end{equation}
where the function $J$ is defined as
\begin{equation}
    J(\tilde{\bfc}, [\tilde{A}], [\tilde{H}], [\tilde{P}]) = \sum_{j=1}^k\left\|\tilde{\bfc}+[\tilde{A}] \hat{{\bfs}}_j+[\tilde{H}]\left(\hat{{\bfs}}_j \otimes \hat{{\bfs}}_j\right)+[\tilde{P}] \hat{{\bfg}}\left(\hat{{\bfs}}_j\right)-\frac{\mathrm{d} \hat{{\bfs}}_j}{\mathrm{~d} t}\right\|_2^2\,.
\end{equation}
The nonnegative scalars $\{\lambda_i\}_{i = 1}^3$ are Tikhonov regularization parameters that are introduced to promote stability in the inferred operator and prevent overfitting \cite{mcquarrieDatadrivenReducedorderModels2021a}. This measure enhances the capability of the ROM for issuing reliable predictions outside the regime in which the data were acquired. The time derivative of the reduced state vector can be estimated numerically, using finite differences for instance.

\begin{remark}
The use of a neural network model is another natural choice to define the nonlinear enrichment function $\mathcal{N}$ (see, e.g., \cite{BARNETT2023112420}), owing to the well-proven nonlinear approximation capabilities of such models:
\begin{equation}
    \mathcal{N}(\hat{\bfs}(t)) = \text{NN}(\hat{\bfs}(t))\,.
\end{equation}
In this case, the reduced-order state-space variable corresponds to the standard POD projection of its physical counterpart, $\hat{\bfs}(t) = [V_r]^\top({\bfs}(t) - {\bfs}_{\text{ref}})$. The reduced-order model associated with this representation can be obtained through a Galerkin projection procedure. The implementation of such a framework is intrusive and complex. As such, this model is not considered in this work and the deployment of the proposed probabilistic formulation in this context is left for future work.
\label{rem: remarkNNIntrusive}
\end{remark}

\section{Probabilistic Formulation for Model-Form Uncertainties}
\label{subsec:SROM}

The goal of this section is to introduce a formulation that enables the description and propagation of model-form uncertainties, using probability theory. Section \ref{subsec:stochastic_representation} introduces a stochastic representation of the projection matrix, laying the groundwork for the probabilistic framework from Section \ref{subsec:Parameterization}. We then outline in Section \ref{subsec: SelectionOperator} how these model-form uncertainties are propagated through the stochastic reduced-order model. The proposed modeling approach is explained step-by-step in Section \ref{subsec:summary}.

\subsection{Stochastic Representation of the Projection Matrix}
\label{subsec:stochastic_representation}

\subsubsection{Preliminaries}\label{subsubsec:StoMod-Prelim}

Operators, whether they be physics-based or learned, are not amenable to direct randomization if one is concerned with model-form uncertainties---which cannot be modeled using stochastic versions of reduced-order matrix operators. Noticing that a POD projection matrix encodes the action of the forward operator, Soize and Farhat proposed a so-called non-parametric formulation for model-form uncertainties where the projection matrix (here, a reduced-order basis) is made random and fed into a Galerkin-type reduced-order model, involving the deterministic operator, to induce statistical fluctuations in model space \cite{soize2017probabilistic,soize2019probabilistic}. Inspired by this work, Zhang and Guilleminot proposed an alternative probabilistic formulation relevant to a multi-model setting where a family of model proposals---identified with the ensemble of models obtained with different training strategies for example---is assumed  available \cite{zhangRiemannianStochasticRepresentation2023a}. 

To extend this construction to the present context, consider the randomization of the projection matrix $[[V_r],[\overline{V}]]$ in the reduced-order approximation \eqref{eqn: nonlinear_rep}, introduced in Section~ \ref{subsec:PrototypicalForm}. Introduce the pair $([\bfV_r], [\overline{\bfV}])$ of matrix-valued random variables, defined on the probability space $(\Theta, \mathcal{T}, P)$ where $\Theta$ is the sample space, $\mathcal{T}$ is a $\sigma$-algebra on $\Theta$, and $P$ is a probability measure on $(\Theta, \mathcal{T})$. The random matrix $[[\bfV_r][\overline{\bfV}]]$ takes values in the subset $\mathbb{S}_{N, (r+q)}$ of the Stiefel manifold $\text{St}(N, (r+q))$, where
\begin{equation}
    \mathbb{S}_{N, (r+q)} = \left\{[M] \in \text{St}(N, (r+q))~|~[B]^\top[M]) = [0_{N_{CD}, (r+q)}]\right\}.
    \label{eqn: stiefelSubset}
\end{equation}
The matrix $[B]^\top$ represents a boundary condition operator that enforces physical constraints on the system. These constraints are inherent to the original full-order model and must be preserved in the reduced-order representation. Specifically, the full-order solution $\bfs$ is assumed to satisfy
\begin{equation}
\left[B\right]^\top \bfs(t) = \left[0_{N_{CD}, N}\right]\,, \quad \forall t \geq 0\,,
\end{equation}
where $N_{CD}$ is the number of (linear) constraints in the physical system. To maintain consistency between the full-order model and the reduced-order model, it follows that the projection matrix satisfies similar constraints, viz.
\begin{equation} \label{eq:constraints-on-ROB}
[B]^\top [[V_r][\overline{V}]] = [0_{N_{CD}, (r+q)}]\,, 
\end{equation}
which implies that $\left[B\right]^\top\left[V\right] = \left[0_{N_{CD}, r}\right]$ and $\left[B\right]^\top[\overline{V}] = \left[0_{N_{CD}, q}\right]$.

To clarify the multi-model setting, consider a parametric inference problem indexed by parameter $\mu$. For simplicity, we assume that $\mu$ belongs to the closed interval $[\underline{\mu}, \overline{\mu}]$. Learning a latent representation and the operator approximation following the approach in Section~\ref{subsec:PrototypicalForm} leads to the estimation of a projection matrix $[[V_r],[\overline{V}]]$ and parameters $\hat{\bfc}$, $[\hat{A}]$, $[\hat{H}]$ and $[\hat{P}]$ defining the learned dynamics, for a given finite set $\mu^{(i)} = \{\underline{\mu}, \ldots, \overline{\mu}\}$ of real numbers in $[\underline{\mu}, \overline{\mu}]$ with cardinality $|\mu^{(i)}|$. The choice of $\mu^{(i)}$ defines a training scenario, which defines a snapshot matrix $[S^{(i)}] \in \mathbb{R}^{N \times k_i}$, where $k_i$ is the number of snapshots for the $i$-th scenario, the associated projection matrix
\begin{equation}
    [\Phi^{(i)}] = [[V_r(\mu^{(i)})] [\overline{V}(\mu^{(i)})]]\,,  
\end{equation}
and the associated parameters $\hat{\bfc}^{(i)}$, $[\hat{A}^{(i)}]$, $[\hat{H}^{(i)}]$ and $[\hat{P}^{(i)}]$ in the learned approximation, where the superscript highlights dependence on $\mu^{(i)}$.

In order to test the robustness of the learning framework with respect to the training strategy, we consider $m$ training scenarios (selected according to the procedure defined in Section \ref{subsubsec:GPM-ROM}) giving rise to a family $\mathcal{A}$ of $m$ projection matrices, $\mathcal{A} = \{[\Phi^{(1)}], \dots, [\Phi^{(m)}]\}$, termed the \textit{anchor points} in the following. For later use, we introduce the following sets: 
\begin{equation}
\mathcal{V}_r = \{[V_r^{(1)}], \ldots, [V_r^{(m)}]\}\,, \quad
\overline{\mathcal{V}} = \left\{[\overline{V}^{(1)}], \ldots, [\overline{V}^{(m)}]\right\}\,, 
\end{equation}
with $[V_r^{(i)}] = [V_r(\mu^{(i)})]$ and $[\overline{V}^{(i)}] = [\overline{V}(\mu^{(i)})]$. We also introduce the \textit{global} projection matrix $[\Phi^*] = [[V_r^*] \ [\overline{V^*}]] \in \mathbb{S}_{N, (r+q)}$, the definition of which is discussed in Section \ref{subsubsec:GPM-ROM}.

Our goals are to derive a probabilistic model for the random matrix $[[\bfV_r][\overline{\bfV}]]$ and to define a strategy to identify an appropriate learned reduced-order operator. These issues are addressed in order in Sections~\ref{subsubsec:ProbRep} and \ref{subsubsec:GPM-ROM}, respectively.

\subsubsection{Probabilistic Representation on a Subset of the Stiefel Manifold}\label{subsubsec:ProbRep}
Following the standard modeling approach on $\text{St}(N, (r+q))$ (e.g., for interpolation with parametric partial differential equations), the probabilistic formulation is developed using the tangent space $T_{[\Phi^*]}\text{St}(N, (r+q))$ to the manifold $\text{St}(N, (r+q))$ at basepoint $[\Phi^*]$ (see, e.g., \cite{Edelman1998} for definition and properties). From a methodological standpoint, this construction first necessitates the choice of projection and retraction operators, written as
\begin{equation}
    P_{[\Phi^*]} : \text{St}(N, (r+q)) \mapsto T_{[\Phi^*]}\text{St}(N, (r+q))
\end{equation}
and 
\begin{equation}
    R_{[\Phi^*]} : T_{[\Phi^*]}\text{St}(N, (r+q)) \mapsto \text{St}(N, (r+q))\,,
\end{equation}
respectively (see, e.g., Chapter 5 in \cite{Absil-2007}). While many choices were proposed for these operators in the literature \cite{Absil-2007}, it was shown in \cite{zhangRiemannianStochasticRepresentation2023a} that the use of iterative algebraic approximations to the Riemannian projection and retractions operators, denoted by $\log_{[\Phi^*]}^{\text{St}}$ (Riemannian logarithm) and $\exp_{[\Phi^*]}^{\text{St}}$ (Riemannian exponential) respectively, preserves the action of the linear constraints in the following sense:
\begin{enumerate}
    \item If $[M] \in \mathbb{S}_{N, (r+q)}$, then $P_{[\Phi^*]}([M]) = \log{[\Phi^*]}^{\text{St}}([M]) \in \mathbb{S}_{N, (r+q)}$;
    \item If $[\Delta] \in \mathbb{S}_{N, (r+q)}$, then $R_{[\Phi^*]}([\Delta]) = \exp{[\Phi^*]}^{\text{St}}([\Delta]) \in \mathbb{S}_{N, (r+q)}$;
\end{enumerate}
with $[\Phi^*] \in \mathbb{S}_{N, (r+q)}$ (as assumed in Section \ref{subsubsec:StoMod-Prelim}). Based on this observation, and using the fact that the tangent space is a vector space, it follows that 
\begin{equation}
    R_{[\Phi^*]}\left(\sum_{i=1}^m p_i P_{[\Phi^*]}([\Phi^{(i)}]) \right) \in \mathbb{S}_{N, (r+q)} \subset \text{St}\left(N, (r+q)\right), \quad \forall \mathbf{p} = \left(p_1, \ldots, p_m\right)^\top \in \mathcal{S}_\mathbf{p}\,,
    \label{eqn: retractDelta}
\end{equation}
where the admissible set $\mathcal{S}_\mathbf{p} \subset \mathbb{R}^m$ is such that the Riemannian operators are well-defined. 

Eq.~\eqref{eqn: retractDelta} provides a path towards an admissible randomization of the projection matrix $[\Phi] = [[V_r][\overline{V}]]$. Before proceeding with the definition of the underlying probability measure, it is instructive to indicate some advantages and disadvantages associated with the use of Riemannian operators. Starting with the latter, it should be noticed that the computation of such operators can be intensive for large values of $N$, $r$ and $q$. While the use of the approximation algorithms in \cite{zimmermannManifoldInterpolation2021} facilitate deployment for scenarios of practical interest, further work is necessary to accelerate the sampling task for the above representation. On the other hand, the use of these operators offer two important advantages. First, they facilitate statistical inference and in particular, the prescription of the Fr\'echet mean for the fluctuations on the manifold (see Section \ref{subsec:Parameterization}). Second, imposing that the coefficients of $\mathbf{p}$ define a convex combination on the tangent space $T_{[\Phi^*]}\text{St}(N, (r+q))$, it can be shown (under assumptions defined in \cite{Afsari}) that the random version of Eq.~\eqref{eqn: retractDelta} yields samples located in the convex hull defined by the anchor points---which provides a simple interpretation of the probabilistic model, regardless of the dimensions.

Following the above discussion, let $\mathbf{P} = (P_1, \dots, P_m)^\top$ denote the random vector of coefficients on the tangent space. The stochastic projection matrix is then defined as 
\begin{equation}
    [\mathbf{\Phi}] = \exp_{[\Phi^*]}^{\text{St}}\left\{ \sum_{i=1}^m P_i \log_{[\Phi^*]}^{\text{St}}([\Phi^{(i)}])\right\}\,, \quad \mathbf{P} \sim \pi\,.
\end{equation}
The underlying probability measure $\pi$, which implicitly defines the stochastic model on the Stiefel manifold, is derived in the context of information theory \cite{Jaynes1957a,Jaynes1957b}, using the principle of maximum entropy \cite{Shannon1948}. Following the above discussion and leveraging the results derived in \cite{Afsari}, we introduce repulsive constraints defining a convex combination on the tangent space:
\begin{equation}
    \mathbb{E}\{\log(P_i)\} < + \infty\,, \quad 1 \leq i \leq m-1\,, 
\end{equation}
and
\begin{equation}
    \mathbb{E}\{\log(1-\sum_{i = 1}^{m-1} P_i)\} < + \infty\,,
\end{equation}
where $\mathbb{E}$ is the operator of mathematical expectation. These constraints lead to the consideration of a Dirichlet distribution with concentration parameters collected in $\boldsymbol{\alpha}$, $\pi = \mathcal{D}(\boldsymbol{\alpha})$. Note that the probability measure defining the matrix-valued random variable $[\mathbf{\Phi}]$, while not determined explicitly due to the nonlinearity of the mappings involved, is fully determined by the measure $\pi$ on the tangent space, as well as the pullback action of the Riemannian retraction operator.

The above probabilistic construction ensures admissibility in the almost sure sense, meaning that the samples of $[\mathbf{\Phi}] = [[\bfV_r] [\overline{\bfV}]]$ belong to the admissible set $\mathbb{S}_{N, (r+q)}$ and satisfy Eq.~\eqref{eq:constraints-on-ROB}. 

\subsubsection{On Constructing the Global Projection Matrix}
\label{subsubsec:GPM-ROM}
In order to deploy the probabilistic framework and construct the global projection matrix, a set of anchor points $\mathcal{A} = \{[\Phi^{(i)}], \dots, [\Phi^{(m)}]\}$ needs to be selected. This task can be completed in different ways, depending on the ultimate goal(s) of the analysis. 

To capture the uncertainties raised by the training strategy for instance, one can consider $m$ combinations of snapshots corresponding to various initial conditions. In the OpInf framework where reduced-order operators are learned from data, it is necessary to identify these combinations such that the respective operators do not deviate significantly from each other---hence maintaining operator consistency across the anchor points. To achieve this, denote any of the matrix-valued parameters $\hat{\bfc}$, $[\hat{A}]$, $[\hat{H}]$ and $[\hat{P}]$, as $[O]$, and let $[D^{[O]}]$ be the pairwise Frobenius norm distance matrix for this operator, obtained over $n_c$ combinations of parameters. Specifically, we define the entries of $[D^{[O]}]$ as
\begin{equation}
D^{[O]}_{ij} = ||[O^{(i)}] - [O^{(j)}]||_F\,, \quad 1 \leq i,j \leq n_c\,,
\label{eqn: pairwiseFrobenius}
\end{equation}
and introduce its normalized counterpart $[\widetilde{D}^{[O]}]$:
\begin{equation}
\widetilde{D}^{[O]}_{ij} = \frac{D^{[O]}_{ij}}{\text{max}([D^{[O]}]) - \text{min}([D^{[O]}])}\,, \quad 1 \leq i,j \leq n_c\,,
\label{eqn: normalizedFrobenius}
\end{equation}
where the minimum and maximum are taken over all components. 

In order to ensure sufficient variability over the set of anchor points, we first perform Riemannian K-means clustering \cite{zhangKmeansPrincipalGeodesic2020} on the set of all matrices $[\Phi^{(i)}]$ for $i = 1, \dots, n_c$ to identify $m \geq 3$ clusters. The clustering is performed on the subset of the Stiefel manifold $\mathbb{S}_{N, (r+q)} \subset \text{St}(N, (r+q))$. We utilize the iterative algorithm proposed in \cite{giovanisPolynomialChaosExpansions2024} to identify the optimal number of clusters $n_c$. For each cluster, we select as the anchor point $[\Phi^{(i)}] \in \mathcal{A}$ the basis matrix that minimizes the total pairwise distance to other matrices within the same cluster for all reduced-order operators, that is,
\begin{equation}
    \mathcal{A} \ni [\Phi^{(i)}] = \text{arg}\underset{j}{\text{min}} \sum_{k=1}^{N_c} \sum_{[O] \in \mathcal{O}} \widetilde{D}^{[O]}_{jk}\,, \quad 1 \leq i \leq m\,, 
    \quad \mathcal{O} = \{\hat{\bfc}, [\hat{A}], [\hat{H}], [\hat{P}]\}\,,
    \label{eqn: anchorpoint}
\end{equation}
where $N_c$ is the number of matrices in the cluster, $\mathcal{O}$ is the set of matrix-valued parameters being considered, and $j$ indexes over the matrices in the cluster. We then construct the global projection matrix by performing a truncated SVD on the concatenated snapshots of the corresponding anchor points,
\begin{align}
[S^{(*)}] &= [[S^{(1)}], \dots, [S^{(m)}]\,, \\
&\approx [U]_{r+q} [\Sigma]_{r+q} [W^\top]_{r+q}\,,
\label{eqn: globalROB}
\end{align}
where $[U]_{r+q}$ and $[W]_{r+q}$ are the left and right singular vectors of $[S^{(*)}]$ truncated to the first $r+q$ columns, and $[\Sigma]_{r+q}$ is the diagonal matrix of singular values. The global projection matrix is then given by $[\Phi^*] = [U]_{r+q}$. A pseudo-algorithm for the computation of the global projection matrix is provided in Algorithm~\ref{alg: ConstructGlobalProjectionMatrix}.

\begin{remark}
When Riemannian K-means proves computationally intensive, due to numerous geodesic distance calculations and Riemannian exponential/logarithm operations, an alternative approach consists in performing K-means on the spectral embedding of the matrices $[\Phi^{(i)}]$ for $i = 1, \dots, n_c$, followed by standard K-means clustering on the resulting embedding. This approach offers a computationally efficient alternative while still capturing the underlying structure of the data on the Stiefel manifold.
\label{rem: AnchorPointsKMeans}
\end{remark}

\begin{algorithm}[tbp]
\flushleft
\caption{Construction of the Global Projection Matrix}
\begin{algorithmic}[1]
\Require Projection matrices $[\Phi^{(i)}], \ 1 \leq i \leq n_c$; Reduced-order matrix operators $\{\hat{\bfc}^{(i)}, [\hat{A}^{(i)}], [\hat{H}^{(i)}], [\hat{P}^{(i)}]\}, \ 1 \leq i \leq n_c$ for $n_c$ training scenarios
\Ensure Global projection matrix $[\Phi^*]$, Set of anchor points $\mathcal{A}$

\State Compute pairwise distance matrices $[D^{[O]}]$ for each operator using Eq.~\eqref{eqn: pairwiseFrobenius}
\State Normalize distance matrices $[\widetilde{D}^{[O]}]$ using Eq.~\eqref{eqn: normalizedFrobenius}

\State Perform Riemannian K-means clustering on $[\Phi^{(i)}]$ to identify $m \geq 3$ clusters \algorithmiccomment{Alternative: Use spectral embedding followed by standard K-means (see Remark \ref{rem: AnchorPointsKMeans})}

\For{each cluster}
    \State Select anchor point $[\Phi^{(n)}]$ using Eq.~\eqref{eqn: anchorpoint}
\EndFor

\State Construct set of anchor points $\mathcal{A} = \{[\Phi^{(1)}], \dots, [\Phi^{(m)}]\}$

\State Concatenate snapshots from anchor points: $[S^{(*)}] = [[S^{(1)}], \dots, [S^{(m)}]$

\State Perform truncated SVD on $[S^{(*)}]$: $[S^{(*)}] \approx [U]_{r+q} [\Sigma]_{r+q} [W^\top]_{r+q}$

\State Set global projection matrix $[\Phi^*] = [U]_{r+q}$\\
\Return $[\Phi^*]$, $\mathcal{A}$

\end{algorithmic}
\label{alg: ConstructGlobalProjectionMatrix}
\end{algorithm}

\subsection{Parameterization of the Probabilistic Model}
\label{subsec:Parameterization}

The global projection matrix $[\Phi^*]$ serves as a crucial reference point in our probabilistic framework. In order to localize the statistical fluctuations on the manifold, it is desirable to prescribe the Fr\'{e}chet mean over the generated samples. A natural choice is to enforce
\begin{equation}
    \mathbb{E}_\pi\{[\mathbf{\Phi}]\} \approx [\Phi^*]\,.
    \label{eqn: frechetmean}
\end{equation}
This choice is made due to the inherent geometric and statistical properties of $[\Phi^*]$. From a geometric perspective, $[\Phi^*]$ represents an optimal linear subspace that captures the dominant modes of variation across all selected anchor points. Statistically, $[\Phi^*]$ can be interpreted as a form of average or central tendency among the local projection matrices. By construction (definition discussed below), it maximizes the projection of the concatenated data onto a low-dimensional subspace, minimizing the overall reconstruction error in the Frobenius norm sense as evidenced by the Eckart-Young Mirsky theorem \cite{Eckart1936}. 

It was shown in \cite{zhangRiemannianStochasticRepresentation2023a} that this can be achieved by defining the vector of concentration parameters $\boldsymbol{\alpha}$ as 
\begin{equation}
    \boldsymbol{\alpha} = \text{arg}\underset{\boldsymbol{\alpha}\,\in\, \mathbb{R}^{m}_{>0}}{\text{min}} \boldsymbol{\alpha}^\top [Q] \boldsymbol{\alpha}\,,
    \label{eqn: quadprog}
\end{equation}
where $[Q] \in \mathbb{R}^{m \times m}$ is the symmetric positive-definite matrix, the entries of which depend on the anchor points through
\begin{equation}
    Q_{i j}=\operatorname{tr}\left\{\log _{[\Phi^*]}^{\text{St}}\left([\Phi^{(i)}]\right)^\top \log _{[\Phi^*]}^{\text{St}}\left([\Phi^{(j)}]\right)\right\}\,,    
\end{equation}
with ``$\operatorname{tr}$'' the trace operator. In practice, the quadratic programming problem defined by Eq.~\eqref{eqn: quadprog} is solved once, prior to any uncertainty quantification analysis. In this study, the Python package \texttt{cvxpy} is used for this purpose.

\subsection{Selection of Reduced-Order Matrix Operators for Uncertainty Propagation}
\label{subsec: SelectionOperator}

In order to propagate model-form uncertainties, Monte Carlo simulations are used in combination with the stochastic reduced-order model. Note that other non-intrusive stochastic solvers, such as stochastic collocation methods, can be used; see \cite{ghanem2017handbook} for a review. Observe that while all reference ``models'' (encoded through the projection matrices $[\Phi^{(i)}]$, $1 \leq i \leq n_c$), obtained by running over all possible combinations of training parameters, are associated with their corresponding reduced-order matrix operators, the new samples for the projection matrix (i.e., the $j$th sample $[\Phi(\theta_j)]$ of $[\mathbf{\Phi}]$) are not associated with any such matrices. Hence, the approach requires the identification of appropriate reduced-order matrix operators for a given sample $[\Phi(\theta_j)]$, so as to minimize bias during time integration in the reduced space. We proceed to introduce a criterion for the selection of suitable reduced-order matrix operators. 

Here, we employ a probabilistic interpretation leveraging the convex combination definition of the samples. Specifically, for each sample $[\Phi(\theta_j)]$, we define an index $\mathcal{I}_j$, with $1 \leq \mathcal{I}_j \leq m$, such that:
\begin{equation}
    P_{\mathcal{I}_j}(\theta_j) = \max \{ P_1(\theta_j), \dots, P_{m}(\theta_j) \}\,.
\end{equation}
This criterion associates $[\Phi(\theta_j)]$ with the closest anchor point $[\Phi^{(\mathcal{I}_j)}]$, operator-wise. Time integration in the reduced space is then carried out with the selected reduced-order matrix operators $\hat{\bfc}^{(\mathcal{I}_j)}$, $[\hat{A}^{(\mathcal{I}_j)}]$, $[\hat{H}^{(\mathcal{I}_j)}]$, and $[\hat{P}^{(\mathcal{I}_j)}]$ for the $j$th sample.

\begin{algorithm}[tbp]
\flushleft
\caption{Stochastic Reduced-Order Modeling with Model-Form Uncertainties}
\begin{algorithmic}[1]
\Require Snapshot matrices $[S^{(i)}]$ for $1 \leq i \leq n_c$ training scenarios
\Ensure Stochastic projection matrix $[\mathbf{\Phi}]$; Reduced-Order Matrix Operators $\{\hat{\bfc}^{(n)}$, $[\hat{A}^{(n)}]$, $[\hat{H}^{(n)}]$, $[\hat{P}^{(n)}]\}, \ 1 \leq n \leq m$ for $m$ anchor points
\For{each training scenario $i$}
    \State Compute POD basis $[\Phi^{(i)}]$ using SVD on $[S^{(i)}]$
    \State Infer reduced-order matrix operators $\hat{\bfc}^{(i)}$, $[\hat{A}^{(i)}]$, $[\hat{H}^{(i)}]$, $[\hat{P}^{(i)}]$ using \eqref{eqn: OpInfLeastSquares}
\EndFor
\State $[\Phi^*], \mathcal{A} \gets$ Call to Algorithm \ref{alg: ConstructGlobalProjectionMatrix}
\State Compute $Q_{nl} = \text{tr}\{\log_{[\Phi^*]}^{\text{St}}([\Phi^{(n)}])^\top \log_{[\Phi^*]}^{\text{St}}([\Phi^{(l)}])\}$
\State Solve $\boldsymbol{\alpha} = \arg\min_{\boldsymbol{\alpha} \in \mathbb{R}^{m}_{>0}} \boldsymbol{\alpha}^\top [Q] \boldsymbol{\alpha}$
\State Generate $\mathbf{P} \sim \mathcal{D}(\boldsymbol{\alpha})$
\State Compute $[\mathbf{\Phi}] = \exp_{[\Phi^*]}^{\text{St}}\{ \sum_{n=1}^{m} P_n \log_{[\Phi^*]}^{\text{St}}([\Phi^{(n)}])\}$
\For{each sample $j$}
    \State Select $\mathcal{I}_j = \arg\max_n P_n(\theta_j)$
    \State Use $[\Phi(\theta_j)], \hat{\bfc}^{(\mathcal{I}_j)}, [\hat{A}^{(\mathcal{I}_j)}], [\hat{H}^{(\mathcal{I}_j)}], [\hat{P}^{(\mathcal{I}_j)}]$ for propagation
\EndFor \\
\Return $[\mathbf{\Phi}]$, $\{\hat{\bfc}^{(n)}$, $[\hat{A}^{(n)}]$, $[\hat{H}^{(n)}]$, $[\hat{P}^{(n)}]\}$
\end{algorithmic}
\label{alg: SummaryAlgorithm}
\end{algorithm}

\subsection{Summary of the Proposed Approach}
\label{subsec:summary}

The proposed stochastic reduced-order modeling approach begins with the generation of $n_c$ snapshot matrices, followed by the completion of four subsequent tasks. First, projection matrices $[\Phi^{(i)}]$ and reduced-order matrix operators $\hat{\bfc}^{(i)}$, $[\hat{A}^{(i)}]$, $[\hat{H}^{(i)}]$, $[\hat{P}^{(i)}]$ are computed for $1 \leq i \leq n_c$ training scenarios. Second, $m$ anchor points are identified via clustering techniques, followed by the construction of a global projection matrix. Third, stochastic samples of the projection matrix are generated through a convex combination of anchor points on the tangent space, employing a Dirichlet distribution for the coefficients. Finally, appropriate reduced-order matrix operators are selected for uncertainty propagation based on a probabilistic interpretation of the stochastic samples. This approach is summarized in (pseudo-)Algorithm \ref{alg: SummaryAlgorithm}.

\section{Numerical Results}
\label{sec:numerical_results}

This section presents the application of the proposed stochastic reduced-order modeling approach to two fundamental problems in fluid dynamics, namely the Burgers' equation (Section \ref{subsec:burgers}) and the two-dimensional Navier-Stokes equations (Section \ref{subsec:ns}). We evaluate the effectiveness of our method in quantifying and propagating model-form uncertainties in the context of operator inference.

\subsection{Application to Burgers' Equation}
\label{subsec:burgers}

In this section, we present the numerical results obtained from applying our methodology to the Burgers' equation. We begin with a description of the problem, followed by a parametric analysis of the reduced-order model, and conclude with uncertainty quantification results.

\subsubsection{Problem Description}\label{subsec: problemdescription}
Consider the parameterized viscous Burgers' equation
\begin{equation}
    \frac{\partial s}{\partial t}+s \frac{\partial s}{\partial x}  = \frac{1}{\text{Re}}\left(\frac{\partial^2 s}{\partial x^2}\right)\,, \quad x \in \Omega = [0, 1]\,, \quad t \in [0, 8]\,, 
    \label{eqn: burgers}
\end{equation}
where $s: \Omega \times T \mapsto \mathbb{R}$ is the scalar-valued time-dependent velocity and $\text{Re}$ is the Reynolds number. The above equation is supplemented with the initial condition
\begin{equation}
    \begin{aligned}
    s(x, 0 ; \mu) = \begin{cases}\mu \sin (2 \pi x) & \text { if } x \in[0,0.5]\,, \\ 0 & \text { otherwise\,,}\end{cases}
    \end{aligned}  
\end{equation}
and zero boundary conditions on the ends of $\Omega$. In this work, we consider $\operatorname{Re} = 1,000$, which leads to sharp gradients in the solution and results in a problem where the Kolmogorov N-width decreases much slower than the commonly considered $\operatorname{Re} = 100$ in the operator learning literature. This choice of Reynolds number is deliberate, as it creates a more challenging scenario for reduced-order modeling. 

To collect snapshots for training the operators, we consider the range $t\in[0, 2]$ for training data, and discretize the spatial domain $\Omega$ with $256$ $\text{P}_1$ Lagrange elements. The step size used for time integration is $\Delta t = 10^{-3}$ and a total of $k=2001$ snapshots are collected for each value of $\mu$ for training the operators. We rely on implicit backward Euler scheme in conjunction with a Newton-Raphson solver for time integration. We collected snapshots for different initial condition parameter $\mu = [0.4, 0.5, \dots, 1.2]$ with an interval of 0.1 and subsequently constructed multiple training datasets based on combinations of parameter $\mu$.
\begin{figure}[tbp]
    \centering
    \includegraphics[width = \textwidth]{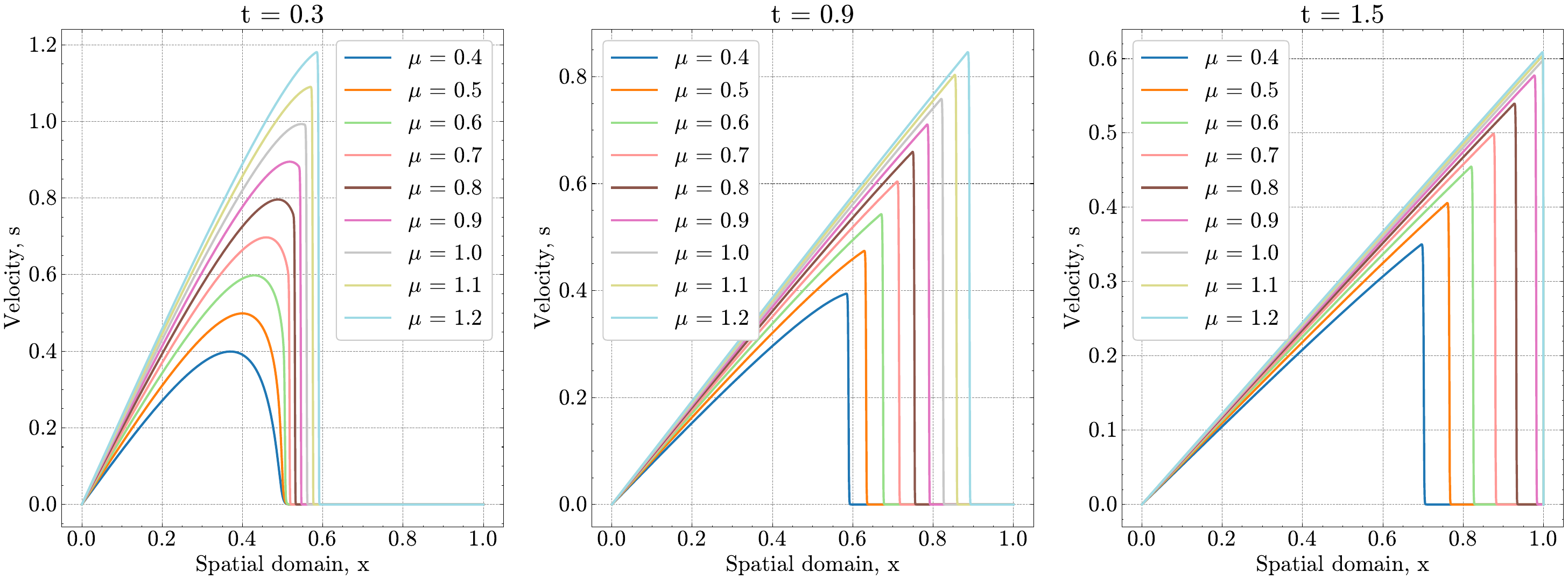}
    \caption{Full-order solution of the Burgers' equation \eqref{eqn: burgers} at various time instances for different values of the initial condition parameter $\mu$. The horizontal axis ($x \in \Omega$) represents the spatial domain, while the vertical axis ($s$) denotes the scalar-valued time-dependent velocity.}
    \label{fig: FullOrderSolution}
\end{figure}
The full-order solution of the Burgers' equation \eqref{eqn: burgers} at various time instances for different initial condition parameter $\mu$ is shown in Fig.~\ref{fig: FullOrderSolution}. The testing window employed for this problem is $t\in[2, 8]$.

\subsubsection{Parametric Analysis on Reduced-Order Model}
\label{subsec: parametricanalysis}
We now turn our attention to the parametric analysis of the ROM obtained via the OpInf method. The reduced-order matrix operators $\hat{\bfc}$, $[\hat{A}]$, $[\hat{H}]$, and $[\hat{P}]$ are inferred using the OpInf method for learning low-dimensional dynamical systems. For the training dataset, we consider all $\mu$ values. We define the reference state $s_{\text{ref}}$ as the mean over all snapshots in the training dataset, and the snapshot matrix $[S]$ is mean-centered. The regularization parameters in the OpInf procedure are determined via grid search, optimizing for minimal relative state error across the training dataset \cite{mcquarrieDatadrivenReducedorderModels2021a}. To distinguish between different Operator Inference (OpInf) approaches, we introduce the following nomenclature:
\begin{enumerate}
    \item POD-OpInf: This denotes the standard OpInf method using linear Proper Orthogonal Decomposition (POD) without any polynomial enrichment.
    \item MPOD-OpInfPoly: This refers to the OpInf method with POD-based representation enriched by polynomial terms. Here, 'M' stands for 'Manifold', indicating the use of a nonlinear manifold approach.
\end{enumerate}
MPOD-OpInfPoly incorporates nonlinear manifold representations, while POD-OpInf uses a purely linear reduced basis. The 'Poly' suffix indicates the inclusion of polynomial terms in the reduced-order model. The time derivative of the reduced state vector is estimated numerically using a fourth-order central finite difference scheme. We then integrate the reduced-order model using the RK45 Explicit Runge-Kutta method, and obtain the reconstructed solution using the nonlinear representation in Eq.~\eqref{eqn: nonlinear_rep}.

\begin{figure}[tbp]
    \centering   
    \begin{subfigure}[t]{0.48\textwidth}
        \centering \includegraphics[width=0.85\textwidth]{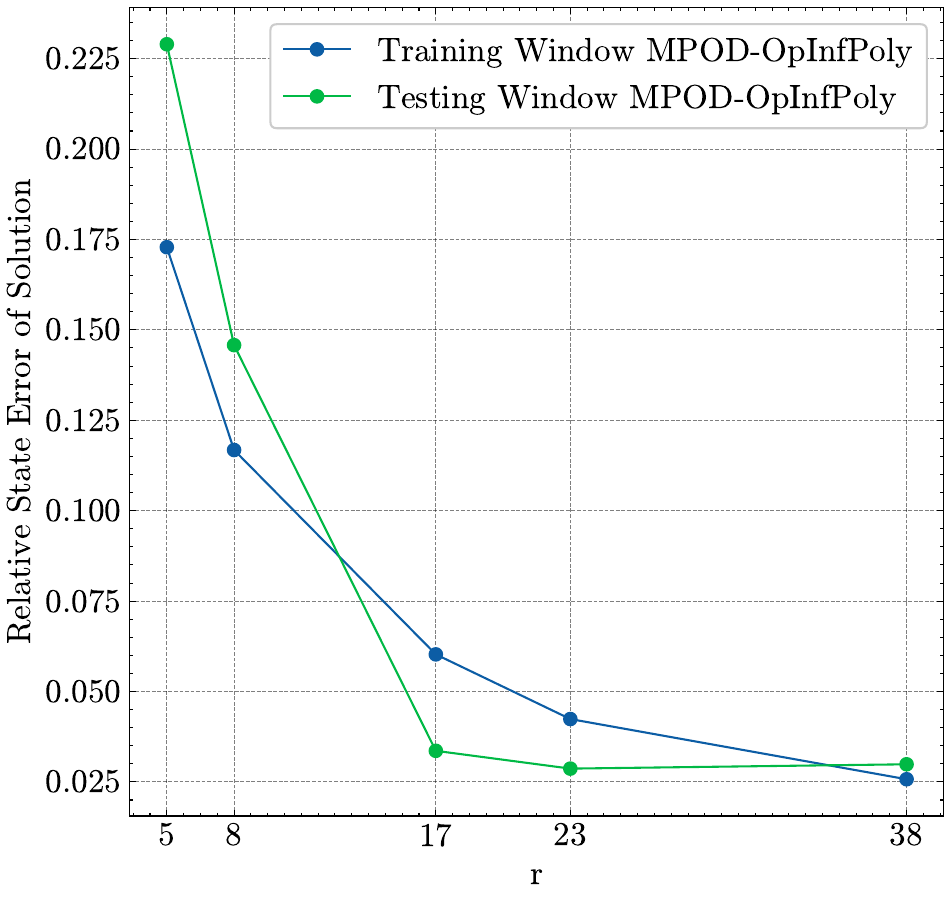}
        \caption{}
        \label{fig: relerr_vs_r_poly}
    \end{subfigure}
    ~
    \begin{subfigure}[t]{0.48\textwidth}
        \centering \includegraphics[width=0.85\textwidth, height=0.8\textwidth]{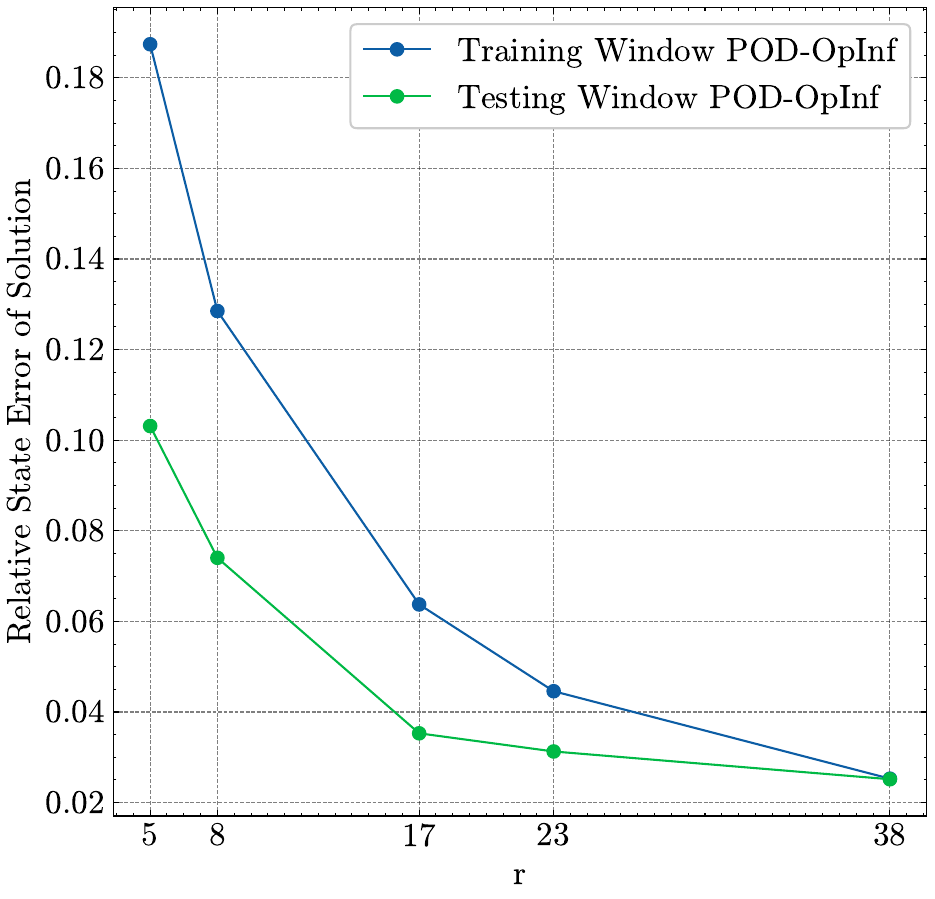}
        \caption{}
        \label{fig: relerr_vs_r}
    \end{subfigure}

    \caption{Parametric analysis of the reduced-order model obtained via the OpInf method for the Burgers' equation \eqref{eqn: burgers}. The horizontal axis represents the number of reduced basis vectors $r$ in $[V]$, which corresponds to $\epsilon(r)$ values of $1\times 10^{-1}$, $5\times 10^{-2}$, $1 \times 10^{-2}$,
    $5 \times 10^{-3}$, and $1 \times 10^{-3}$. (a) Relative state error \eqref{eqn: relativerror} as a function $r$ with $q=2$ basis vectors in $[\overline{V}]$ for MPOD-OpInfPoly. (b) Relative state error as a function of $r$ for POD-OpInf.}
    \label{fig: paramAnalysis_OpInf_r}
\end{figure}

Fig.~\ref{fig: paramAnalysis_OpInf_r} shows the parametric analysis of the reduced-order model obtained via the OpInf method for the Burgers' equation \eqref{eqn: burgers} with varying number of reduced basis vectors $r$ in $[V]$ corresponding to different truncation order based on analyzing the convergence of the error function
\begin{equation}
    \epsilon(r) = 1 - \frac{\sum_{i=1}^{r}\sigma_{i}^2}{\sum_{i=1}^{N}\sigma_{i}^2}\,, \quad \sigma_i = [\Sigma]_{ii}\,,
    \label{eqn: errorfunction}
\end{equation}
with $N = n_x + 1$, by considering the singular value decomposition of the snapshot matrix $[S] = [U] [\Sigma] [W^\top]$. The relative state error is computed as
\begin{equation}
    \frac{||[S] - \Gamma\left([\hat{S}]\right)||_F}{||[S] - [S_{\text{ref}}]||_F}\,.
    \label{eqn: relativerror}
\end{equation}

It is observed that MPOD-OpInfPoly has a better relative state error compared to the POD-OpInf model at lower values of $r$. This is attributed to the polynomial representation of the operators, which allows for a more accurate representation of the dynamics. However, as the number of reduced basis vectors $r$ increases, the relative state error of the POD-OpInf model decreases and eventually converges to the MPOD-OpInfPoly model. This suggests that the polynomial representation of the operators is not necessary when a sufficient number of reduced basis vectors are used. Outside the range of training data ($t \in [2, 8]$), the relative state error of the MPOD-OpInfPoly model is higher than the MPOD-OpInf model at lower values of $r$, which could be attributed to the overfitting of the polynomial representation. These results are consistent with the observations reported in \cite{geelenLearningPhysicsbasedReducedorder2023}. Note that better results are obtained in testing for the POD-OpInf procedure. A possible explanation is that the testing region (from $t=2$ to $t=8$) mainly comprises diffusive patterns which are easily captured by the dominant POD modes, whereas the training region contains shocks or sharper gradients that might require a larger number of POD modes to resolve.

\begin{figure}[tbp]
    \centering
    \includegraphics[width = 0.4\textwidth]{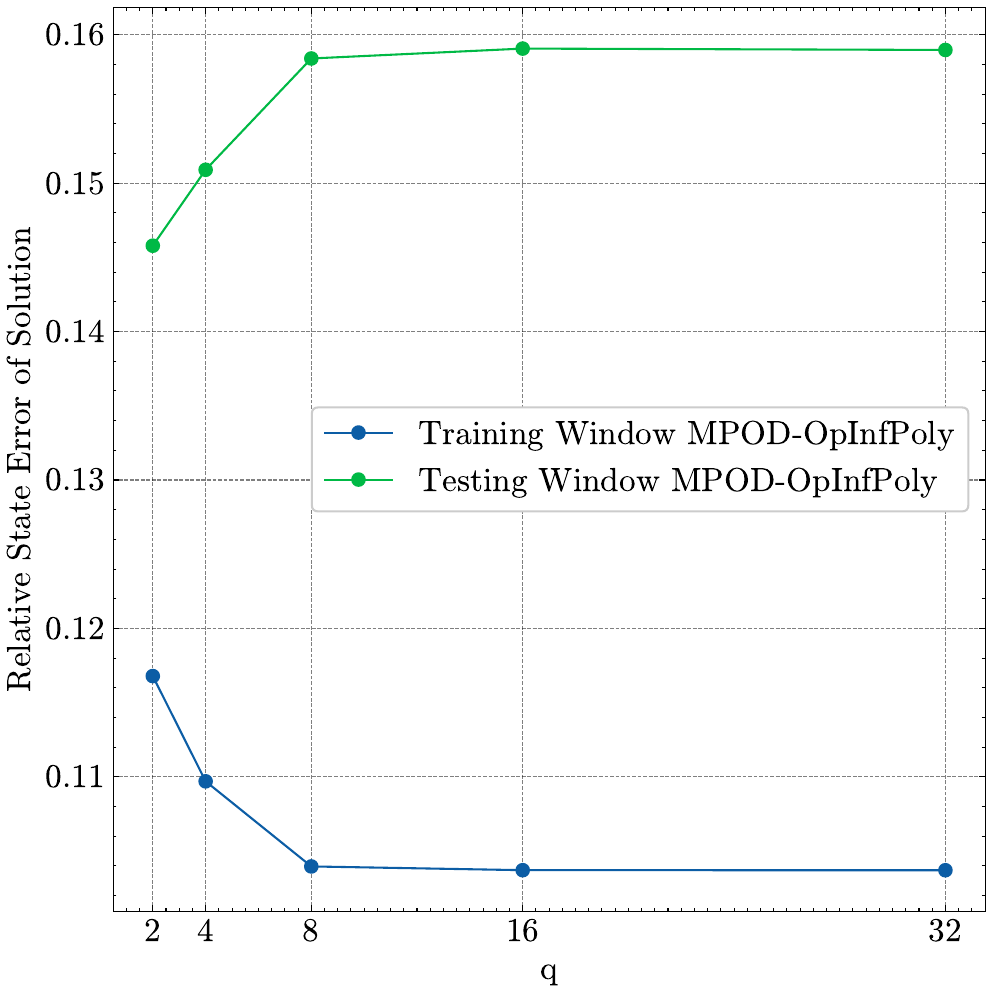}
    \caption{Relative state error \eqref{eqn: relativerror} as a function of the number of reduced basis vectors $q$ in $[\overline{V}]$ for MPOD-OpInf. The horizontal axis denotes $q$ number of reduced basis vectors in $[\overline{V}]$.}
    \label{fig: relerr_vs_q_poly}
\end{figure}

Fig.~\ref{fig: relerr_vs_q_poly} illustrates the relationship between the relative state error and the number of reduced basis vectors $q$ in $[\overline{V}]$ for the MPOD-OpInfPoly model. We observe a notable trend for the MPOD-OpInfPoly model: within the training range, the relative state error decreases as $q$ increases, indicating improved model performance. Conversely, beyond the training range, the relative state error generally increases with $q$. This dichotomy suggests that while MPOD-OpInfPoly exhibits robust interpolation capabilities within the training domain, its extrapolation performance may be limited.
To quantify the effectiveness of increasing $q$ for nonlinear approximations of the form \eqref{eqn: nonlinear_rep}, we compute the cumulative snapshot energy metric
\begin{equation}
\varepsilon(r,q) = \frac{||[V][\hat{S}] + [\overline{V}][\Xi][\bfg(\hat{\bfs}_1) \cdots \bfg(\hat{\bfs}_k)]||_F^2}{||[S] - [S{\text{ref}}]||_F^2}.
\label{eqn: modified_error_function}
\end{equation}
With $r=8$, the linear subspace captures 95\% of the snapshot energy. Augmenting this subspace with an additional $q=8$ basis vectors in $[\overline{V}]$ yields only a marginal 2.1\% increase in captured snapshot energy, suggesting diminishing returns for higher values of $q$.

\subsubsection{Uncertainty Quantification Results}
In this section, the stochastic reduced-order modeling approach is deployed on the Burgers' equation \eqref{eqn: burgers} to capture model-form uncertainties in the reduced-order model. The projection matrix $[\Phi]$ is randomized using the probabilistic formulation introduced in Section \ref{subsec:SROM}. The global projection matrix $[\Phi^*]$ is constructed using the anchor points obtained based on the construction described in Section \ref{subsec:Parameterization}.  
To obtain the reduced-order matrix operators, $r$ has to be first determined to obtain the reduced state vectors $[\hat{S}]$. Based on the parametric analysis in Section \ref{subsec: parametricanalysis}, an error threshold of $5\times 10^{-2}$ is used to allow further enrichment of the reduced basis vectors while maintaining a low relative state error.
\begin{figure}[tbp]
    \centering
    \includegraphics[width=0.6\textwidth]{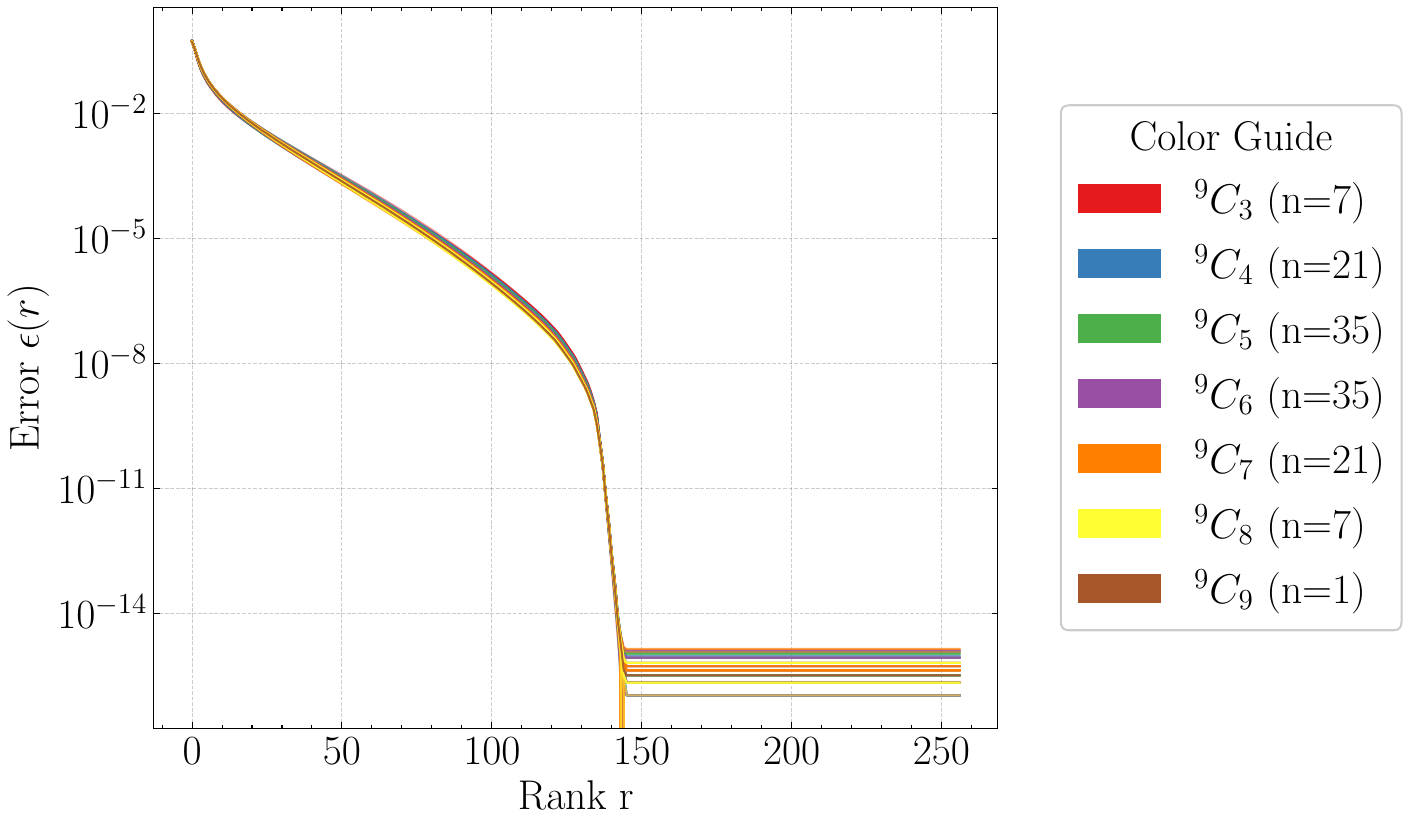}
    \caption{Graph of the error function $r \mapsto \epsilon(r)$ for different combinations of parameter $\mu$. The vertical axis represents the error function $\epsilon(r)$ as defined in Eq.~\eqref{eqn: errorfunction}. The horizontal axis shows the rank $r$ of the projection matrices $[\Phi^{(i)}]$, where $1 \leq i \leq n_c$, and $n_c$ is the number of parameter combinations. Distinct colors indicate different combination groups $^9C_k$ (where $3\leq k \leq 9$ and $^nC_k$ denotes $k$-combinations from a set of $n$ elements). The number of combinations within each group is denoted by $n$ in the figure.}
    \label{fig: errorFunc_all}
\end{figure}
Fig.~\ref{fig: errorFunc_all} shows the graph of the error function $\epsilon(r)$ defined by \eqref{eqn: errorfunction} for all combinations of parameter $\mu$. By choosing the minimum $r$ across all combinations of parameter $\mu$ that satisfies the error threshold, we identify $r=7$. We then infer the reduced-order matrix operators using the OpInf method for $r=7$ and $q=8$.

\begin{figure}[tbp]
    \centering
    \includegraphics[width = 0.6\textwidth]{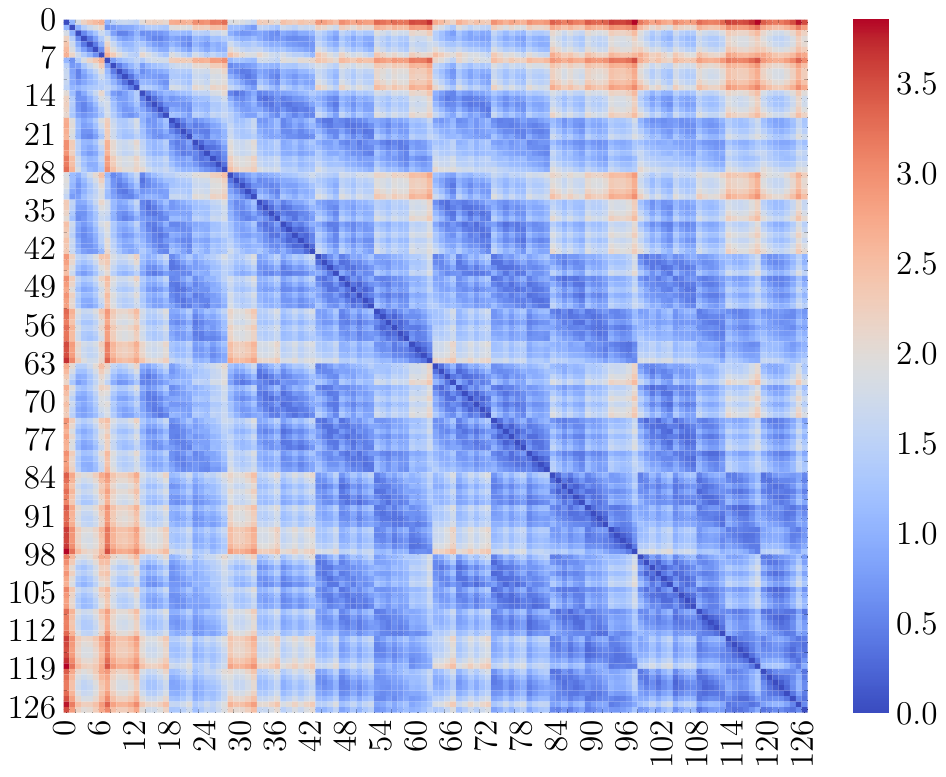}
    \caption{Heat map of the total normalized Frobenius norm difference between the operators for different combinations of snapshots.}
    \label{fig: HM_TotalNormFrobDiff}
\end{figure}

Fig.~\ref{fig: HM_TotalNormFrobDiff} shows the heat map of the total normalized Frobenius norm difference between the operators for different combinations of snapshots. The anchor points are then identified using \eqref{eqn: anchorpoint}.
Three anchor points are identified, which corresponds to $\mu$ values of $\mu^{(1)} = \{0.4, 0.5, 0.7, 0.9, 1.1, 1.2\}$, $\mu^{(2)} = \{0.4, 0.6, 0.8, 0.9, 1.1, 1.2\}$, and $\mu^{(3)} = \{0.4, 0.5, 0.6, 0.9, 1.2\}$.

\begin{figure}[tbp]
    \centering
    \begin{subfigure}{0.49\textwidth}
    \includegraphics[width=\textwidth]{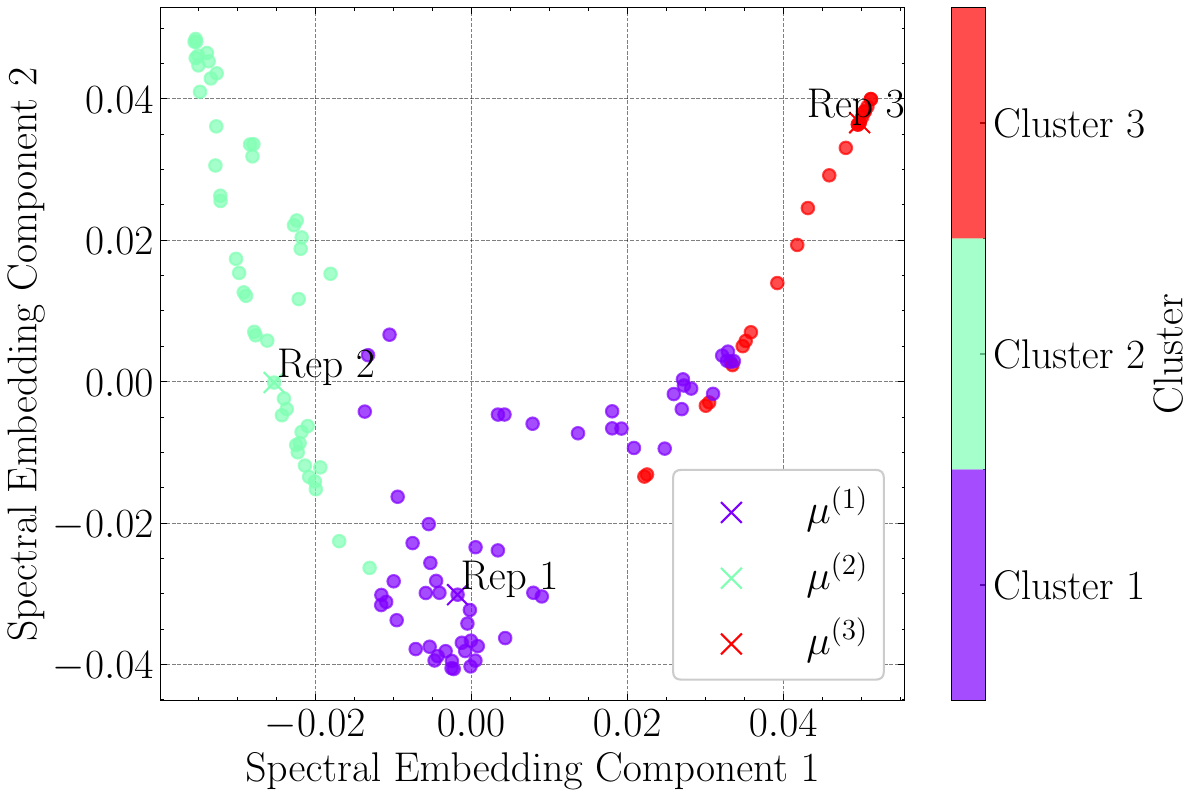}
    \caption{}
    \label{fig: SE_V_basis_RiemK_Cluster}
    \end{subfigure} \hspace{1em}
    \begin{subfigure}{0.45\textwidth}
    \includegraphics[width=\textwidth]{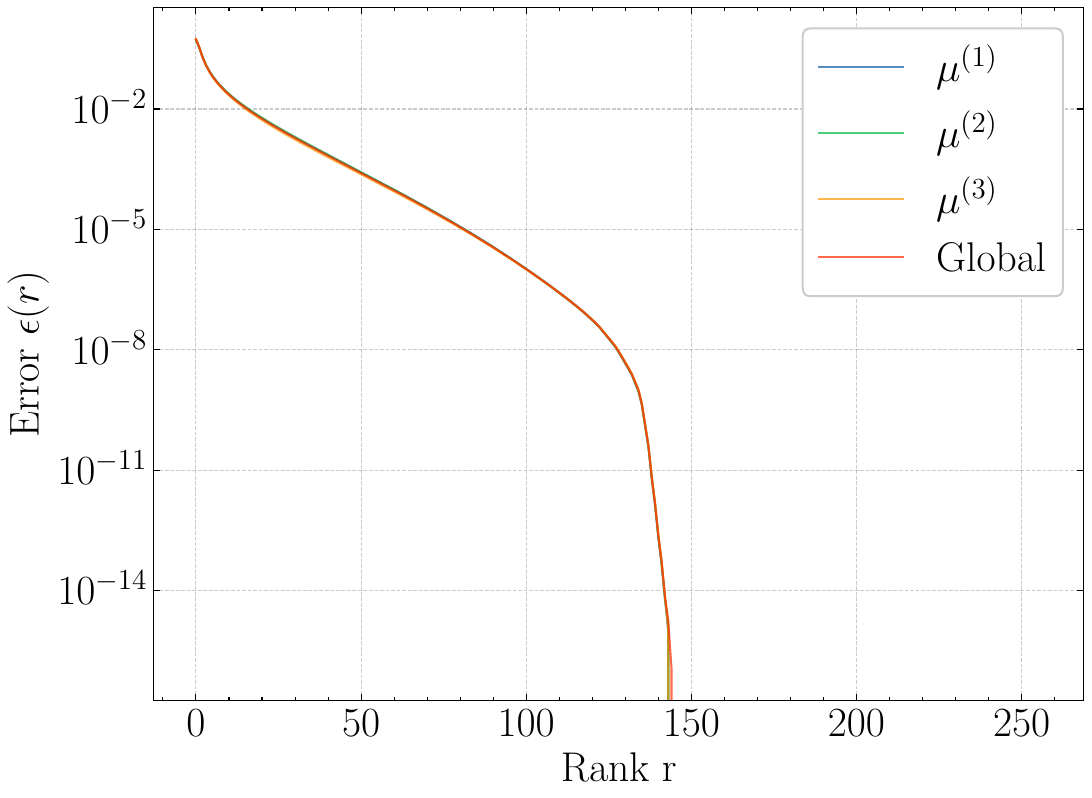}
    \caption{}
    \label{fig: errorFunc_Global_r7}
    \end{subfigure}
    \caption{(a) Spectral embedding of the set of all matrices obtained from combinations of snapshots for different initial condition parameter $\mu^{(i)}\,, 1 \leq i \leq n_c$. The anchor points $\mathcal{A} = \{[\Phi^{(n)}], \ 1 \leq n \leq 3\}$ are annotated and the clusters are color-coded. (b) Graph of the error function $r \mapsto \epsilon(r)$ for the anchor points $\mathcal{A}$ and the global projection matrix $[\Phi^*]$. The vertical axis represents the error function $\epsilon(r)$ as defined in Eq.~\eqref{eqn: errorfunction}. The horizontal axis shows the rank $r$ of the global projection matrix $[\Phi^{*}]$ and the projection matrices $[\Phi^{(n)}] \in \mathcal{A}$, where $1 \leq n \leq 3$.}
\end{figure}

Fig.~\ref{fig: SE_V_basis_RiemK_Cluster} shows the spectral embedding of the set of all matrices obtained from combinations of snapshots for different initial condition parameter $\mu$ using the POD approach. The anchor points are annotated and their respective clusters are color-coded. We then obtain the global projection matrix $[\Phi^*]$ similarly using the POD approach. Fig.~\ref{fig: errorFunc_Global_r7} shows the graph of the error function $\epsilon(r)$ for the anchor points $\mathcal{A}$ and the global projection matrix $[\Phi^*]$. With the same error threshold, $r=7$ is selected to build the anchor points and global projection matrix $[\Phi^*]$, leading to stochastic modeling in $\mathbb{S}_{257, {(7+8)}} \subset \text{St}_{257, {(7+8)}}$.

\begin{figure}
    \centering
    \includegraphics[width=\textwidth]{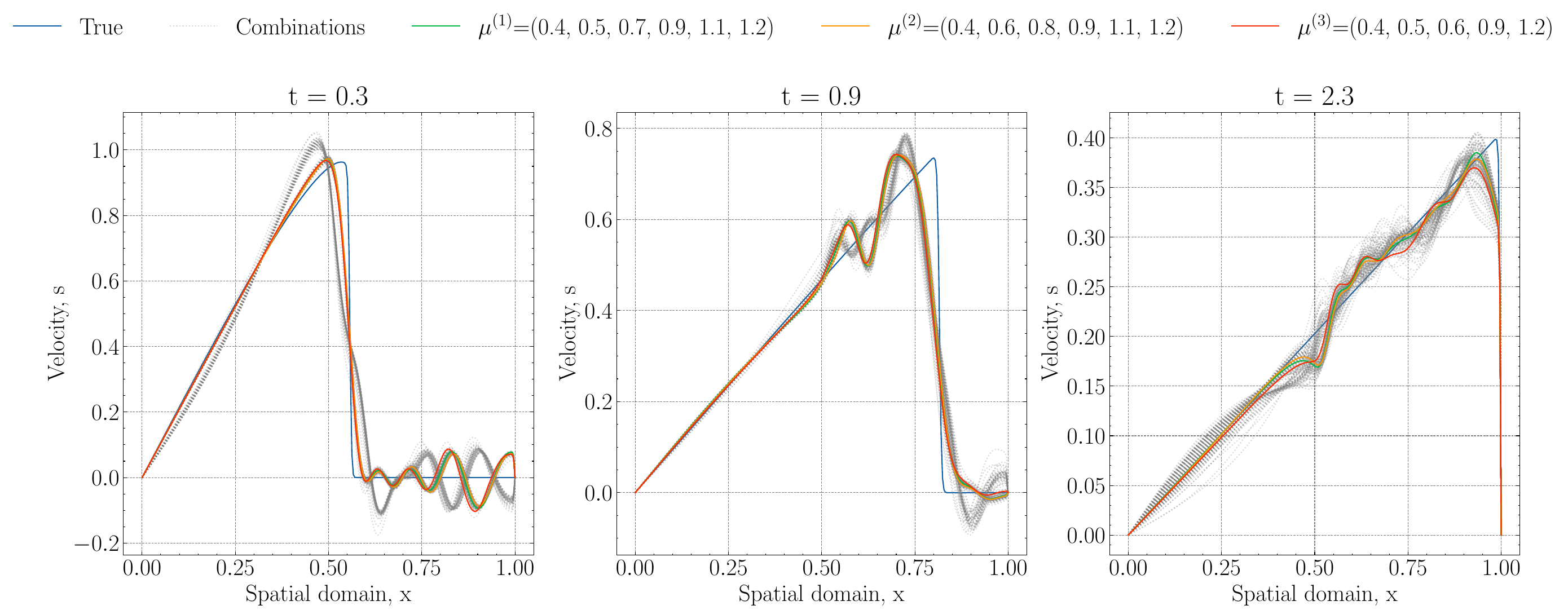}
    \caption{Reconstructed solution of the Burgers' equation \eqref{eqn: burgers} with operators obtained from different combinations of parameter $\mu$. The corresponding reconstructed solution of the anchor points are highlighted.}
    \label{fig: soln_comb_selected_operators_q8_r7}
\end{figure}

Fig.~\ref{fig: soln_comb_selected_operators_q8_r7} presents the reconstructed solution of the Burgers' equation \eqref{eqn: burgers} with operators obtained from different combinations of parameter $\mu$. The corresponding reconstructed solutions of the anchor points are highlighted. We observe that the reconstructed solutions of the anchor points do not deviate significantly from each other, maintaining consistency across the anchor points.

To generate realizations of the stochastic projection matrix $[\mathbf{\Phi}]$, the concentration parameters $\boldsymbol{\alpha}$ are obtained by solving the 
quadratic programming problem \eqref{eqn: quadprog}. It is found that $\boldsymbol{\alpha} = (0.5394, 0.2423, 0.2183)^\top$. The stochastic projection matrix $[\mathbf{\Phi}]$ is then generated using the Dirichlet distribution with concentration parameters $\boldsymbol{\alpha}$. To observe the localization of generated samples, the spectral embedding approach is used to project the generated samples from $\text{St}(257, 15)$ to $\mathbb{R}^{2}$, as shown in Fig.~\ref{fig: SE_V_basis_RiemK_Cluster}. 
\begin{figure}[tbp]
    \centering
    \includegraphics[width=\textwidth]{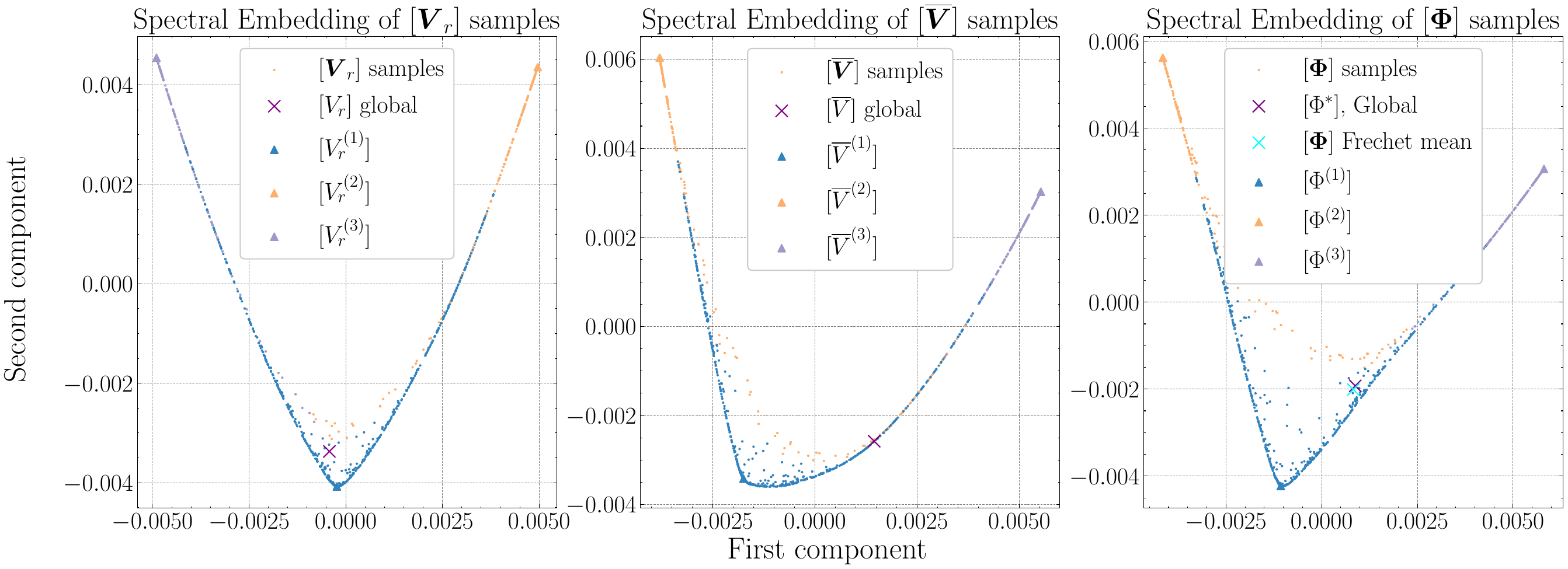}
    \caption{Visualization of the dataset, given by the three anchor points corresponding to $\mu^{(1)}$, $\mu^{(2)}$, and $\mu^{(3)}$, and the 1000 generated samples of the stochastic projection matrix $[\mathbf{\Phi}]$. Spectral embedding is used to project data in $\text{St}(257, 7+8)$ to $\mathbb{R}^{2}$. Left: Spectral embedding visualization of $[\bfV_r]$ Middle: Spectral embedding visualization of $[\overline{\bfV}]$. Right: Spectral embedding visualization of the stochastic projection matrix $[\mathbf{\Phi}]$.}
    \label{fig: SE_StiefelSamples_RiemK_q8_r7}
\end{figure}
The samples are located within the convex hull defined by the anchor points due to the Riemannian convex combination. The Fr\'echet mean of the samples is found to be close to the global projection matrix $[\Phi^*]$, as shown in Fig.~\ref{fig: SE_StiefelSamples_RiemK_q8_r7}, indicating that the constraint on the mean is properly satisfied.

\begin{figure}
    \centering
    \includegraphics[width=\textwidth]{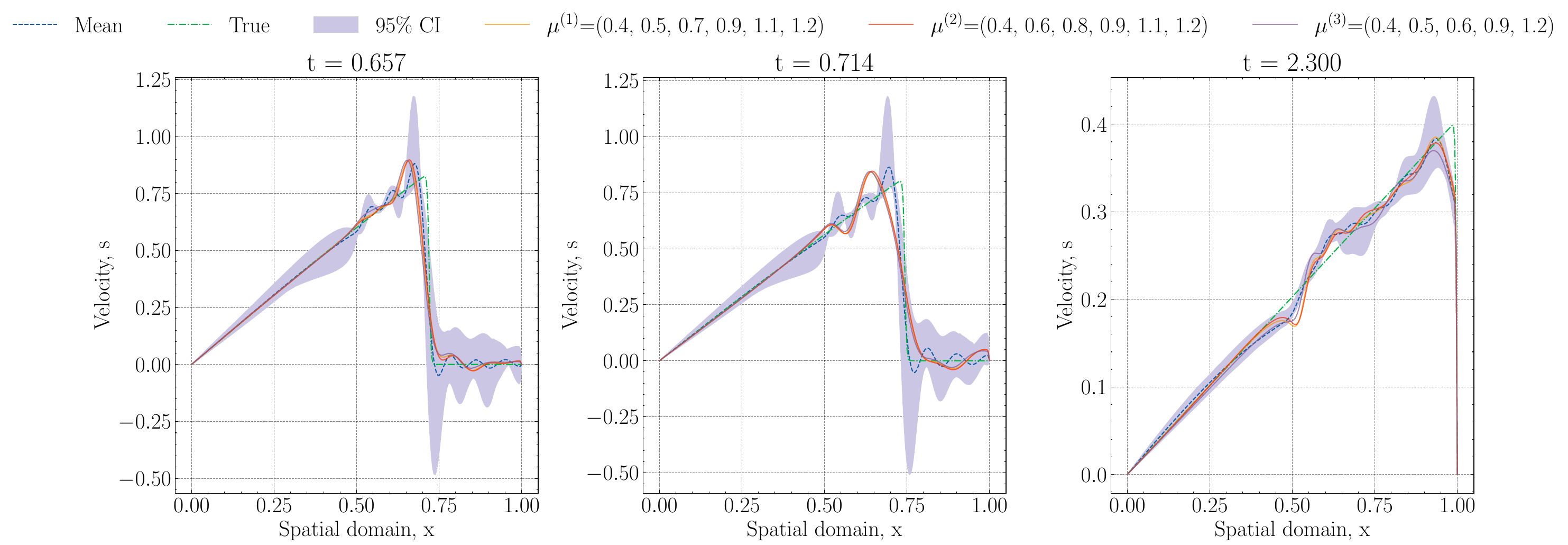}
    \caption{The 95\% confidence interval of the solution of the Burgers' equation \eqref{eqn: burgers} obtained using the stochastic reduced-order model.}
    \label{fig: CI_solution_q8_r7}
\end{figure}

To propagate the model-form uncertainties, we leverage the reduced-order matrix operator selection strategy in Section \ref{subsec: SelectionOperator} and perform time-integration in the reduced space using the Explicit Runge-Kutta method of order 5(4) and the reconstructed solution is obtained using the nonlinear representation \eqref{eqn: nonlinear_rep}. The 95\% confidence interval obtained with the aforementioned strategy is shown in Fig.~\ref{fig: CI_solution_q8_r7}.

\subsection{Application to the two-dimensional Navier-Stokes Equations}
\label{subsec:ns}

In this section, we test our probabilistic methodology to the (canonical) problem of two-dimensional transient flow around a cylinder. We first start with a brief review of the problem in Section \ref{subsec: problemdescription-NS}. We then report on parametric analyses and uncertainty quantification results in Section \ref{subsec: problemdescription-NS} and Section \ref{subsec: UQ-NS}, respectively.

\subsubsection{Problem Description}
\label{subsec: problemdescription-NS}
We consider the incompressible Navier-Stokes equations 
\begin{equation}
    \begin{aligned}
    \rho\left(\frac{\partial \bfu}{\partial t}+\bfu \cdot \nabla \bfu\right) & =\nabla \cdot \sigma(\bfu, p) + \bff\,, \quad t \in [0, 8]\,, \\
    \nabla \cdot \bfu & =0\,, 
    \end{aligned}
    \label{eqn: NS2D}
\end{equation}
where primary variables are the velocity vector $\bfu = \left(u, v\right)^\top$, whose components along the $x$ and $y$-directions are denoted by $u$ and $v$, respectively, and the pressure $p$. Physical parameters are the density $\rho$ and the dynamic viscosity $\mu$. The term $\sigma(\mathbf{u}, p)$ denotes the stress tensor given, for a Newtonian fluid, by
\begin{equation}
\sigma(\bfu, p)=2 \mu \epsilon(\bfu)-p I\,,    
\end{equation}
with $\epsilon(\bfu)$ the strain-rate tensor,
\begin{equation}
    \epsilon(\bfu)=\frac{1}{2}\left(\nabla \bfu+(\nabla \bfu)^\top\right)\,,
\end{equation} 
and $\bff$ is a given force per unit volume. The geometry and parameters are taken from the DFG 2D-3 benchmark in the FeatFlow benchmark suite \cite{FeatFlowBenchmarkSuite}. The problem \eqref{eqn: NS2D} is supplemented with the inlet velocity profile
\begin{equation}
    \bfu(x, y) = \left(\frac{4 U y(0.41-y)}{0.41^2}, 0\right)^\top\,,
\end{equation}
with $U=1.5$. Non-slip boundary conditions, $\bfu = \boldsymbol{0}$, are applied at the walls and around the obstacle. We set $\bff = \boldsymbol{0}$ and $\rho = 1$. To perform the parametric analysis, we consider different values of $\mu$ corresponding to Reynolds numbers equal to 75, 112.5, 135, 150, 165, 187.5, 225, 262.5, 300. 

Regarding data collection, 200 snapshots are retrieved in the time interval $[4, 5]$. Each snapshot is transformed into a column vector with $N = 85,808$ entries. An additional testing dataset with a Reynolds number of 270 was generated for assessing the interpolation capabilities of the operators and to perform uncertainty quantification analysis. The step size used for the time integration is 
$\Delta t = 2 \times 10^{-4}$. We subsequently constructed multiple training datasets based on combinations of the parameter $\mu$. Similar to the Burgers' equation application, we define the reference state $s_{\text{ref}}$ as the mean over all snapshots of a training dataset, and the snapshot matrix $[S]$ is mean-centered for each training dataset.

\subsubsection{Uncertainty Quantification Results}
\label{subsec: UQ-NS}

\begin{figure}[tbp]
    \centering
    \includegraphics[width=0.6\textwidth]{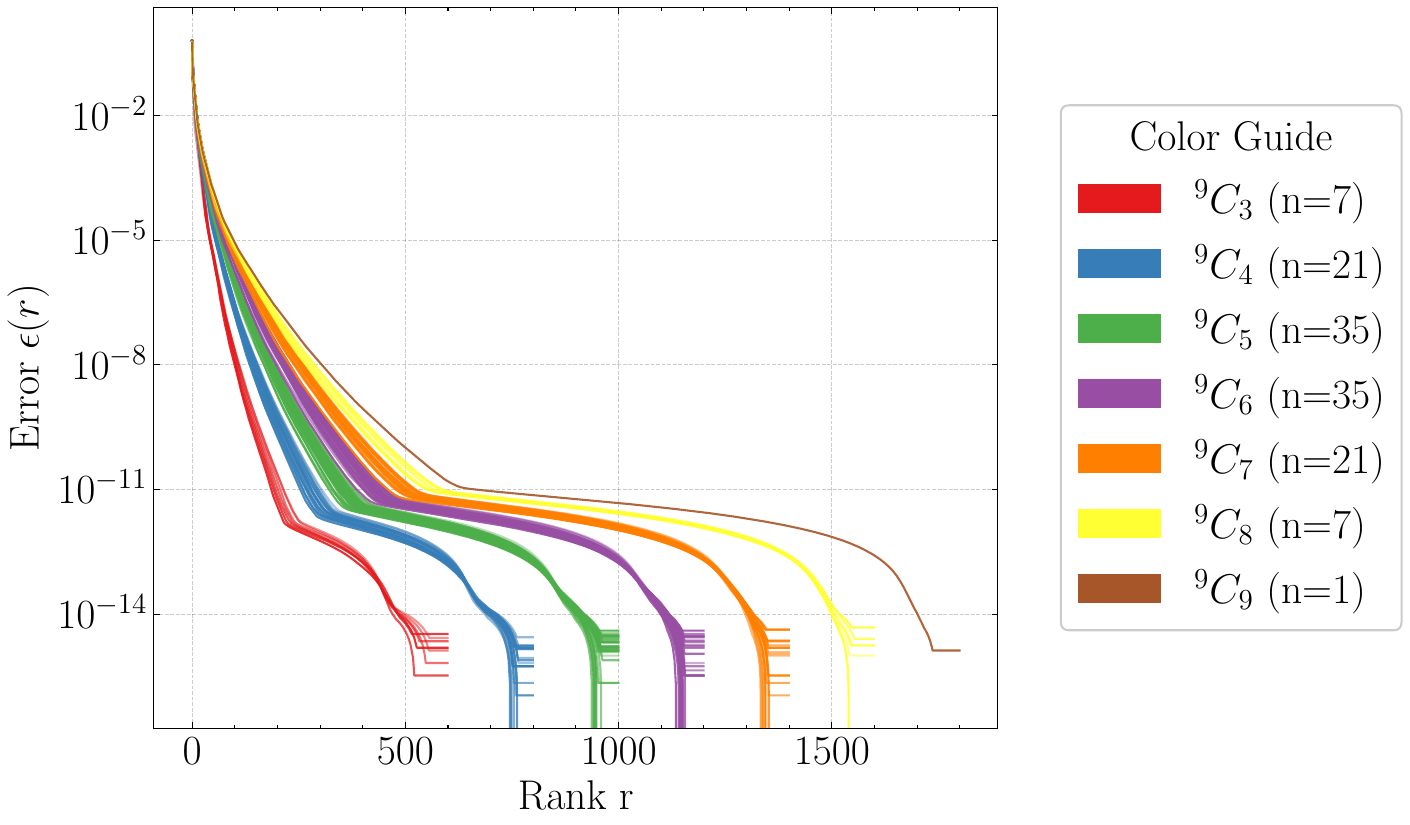}
    \caption{Graph of the error function $r \mapsto \epsilon(r)$ for different combinations of parameter $\mu$ for the two-dimensional transient flow past circular cylinder problem. The horizontal axis shows the rank $r$ of the projection matrices $[\Phi^{(i)}]$, where $1 \leq i \leq n_c$, and $n_c$ is the number of parameter combinations. Distinct colors indicate different combination groups $^9C_k$ (where $3\leq k \leq 9$ and $^nC_k$ denotes $k$-combinations from a set of $n$ elements). The number of combinations within each group is denoted by $n$ in the figure}.
    \label{fig: errorFunc_all_NS}
\end{figure}

The global projection matrix $[\Phi^*]$ is constructed using the anchor points obtained following the construction in Section \ref{subsec:Parameterization}. An error threshold of $5\times 10^{-2}$ is used to allow further enrichment of the reduced basis vectors while maintaining a low relative state error. Fig.~\ref{fig: errorFunc_all_NS} shows the graph of the error function $r \mapsto \epsilon(r)$ defined by \eqref{eqn: errorfunction}, for all values of parameter $\mu$.
By choosing the minimum $r$ across all combinations of parameter $\mu$ that satisfies the error threshold, we identify $r = 3$. With an error threshold of $1 \times 10^{-3}$, 16 POD modes are required. Hence, we choose the remaining $q = 13$ POD modes to be contained in $[\overline{\bfV}]$. The reduced-order matrix operators are then inferred using the OpInf method with $r=3$ and $q=13$. 

Given the increased dimensionality of the matrices and the consequent computational complexity associated with Riemannian K-means, we employ the alternative methodology (Remark~\ref{rem: AnchorPointsKMeans}) proposed in Section \ref{subsec:Parameterization} for anchor point identification. This approach specifically combines spectral embedding with standard K-means clustering, followed by the application of Eq.~\eqref{eqn: anchorpoint} to select representative anchor points within each cluster. The three identified anchor points correspond to the combinations $$\mu^{(1)} = \{\frac{0.5}{75}, \frac{0.15}{150}, \frac{0.15}{165}, \frac{0.15}{262.5}, \frac{0.15}{300}\}\,,$$ $$\mu^{(2)} = \{\frac{0.15}{75}, \frac{0.15}{112.5}, \frac{0.15}{135}, \frac{0.15}{165}, \frac{0.15}{262.5}, \frac{0.15}{300}\}\,,$$ and $$\mu^{(3)} = \{\frac{0.15}{75}, \frac{0.15}{187.5}, \frac{0.15}{225}, \frac{0.15}{300}\}\,.$$ 

\begin{figure}[tbp]
    \centering
    \begin{subfigure}{0.49\textwidth}
    \includegraphics[width=\textwidth]{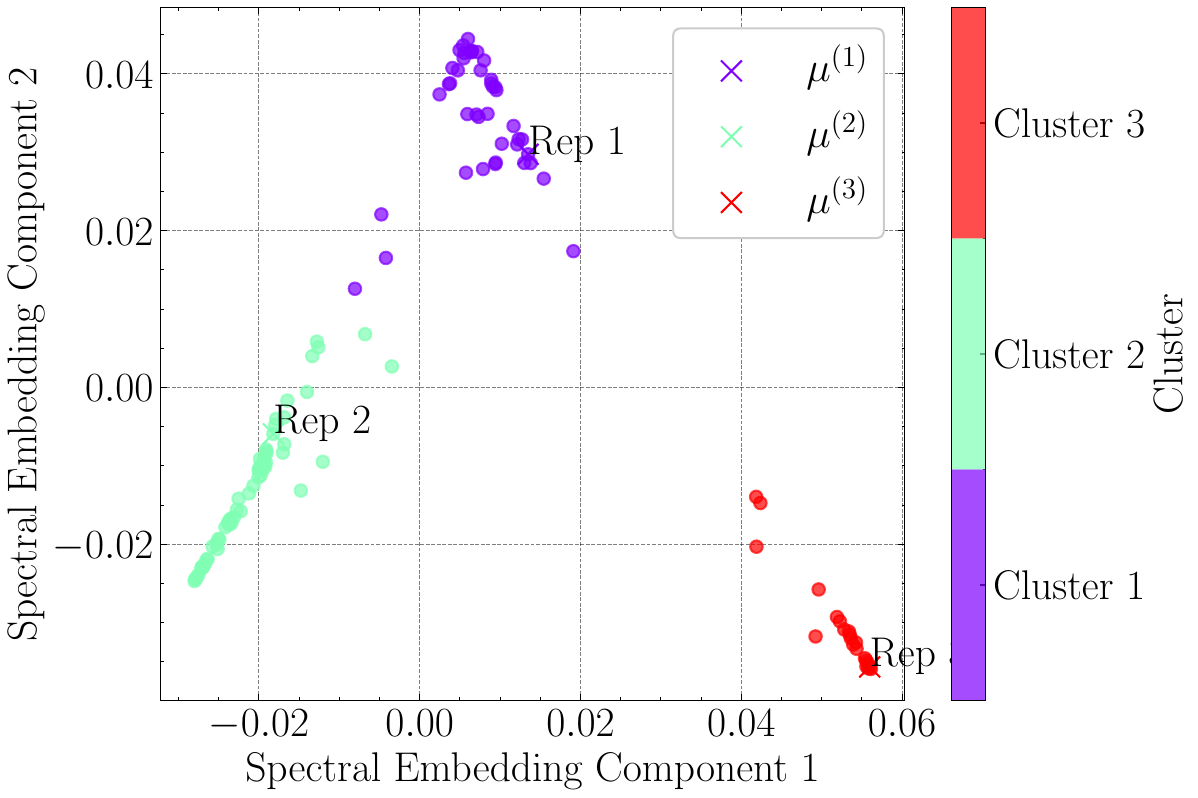}
    \caption{}
    \label{fig: SE_V_basis_KMeans_Cluster_NS_r3}
    \end{subfigure} \hspace{1em}
    \begin{subfigure}{0.45\textwidth}
    \includegraphics[width=\textwidth]{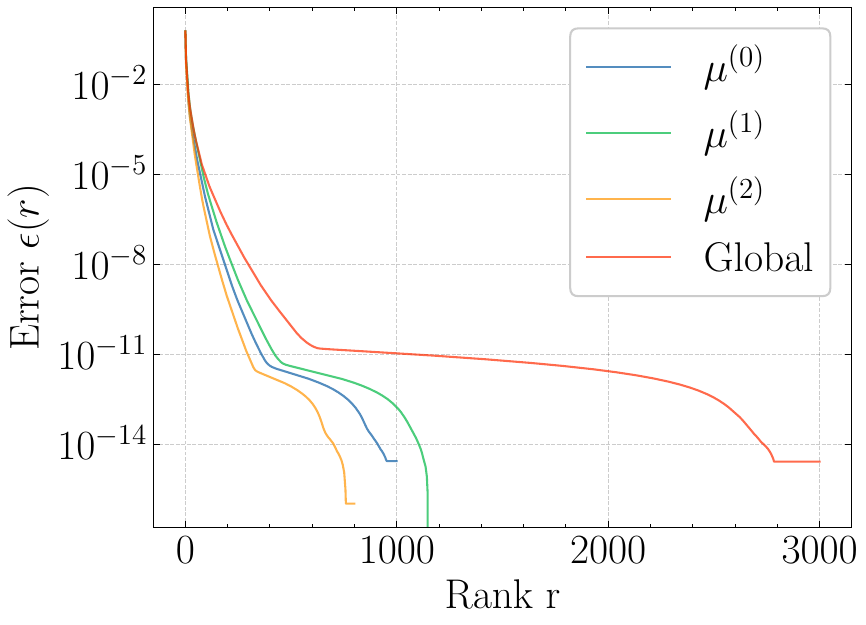}
    \caption{}
    \label{fig: errorFunc_Global_NS_r3}
    \end{subfigure}
    \caption{(a) Spectral embedding of the set of all matrices obtained from combinations of snapshots for different dynamic viscosity parameter $\mu^{(i)}\,, 1 \leq i \leq n_c$. The anchor points $\mathcal{A} = \{[\Phi^{(n)}], \ 1 \leq n \leq 3\}$ are annotated and the clusters are color-coded. (b) Graph of the error function $\epsilon(r)$ for the anchor points $\mathcal{A}$ and the global projection matrix $[\Phi^*]$ for the two-dimensional transient flow past circular cylinder problem. The vertical axis represents the error function $\epsilon(r)$ as defined in Eq.~\eqref{eqn: errorfunction}. The horizontal axis shows the rank $r$ of the global projection matrix $[\Phi^{*}]$ and the matrices $[\Phi^{(n)}] \in \mathcal{A}$, where $1 \leq n \leq 3$.}
\end{figure}

Fig.~\ref{fig: SE_V_basis_KMeans_Cluster_NS_r3} shows the spectral embedding of the set of all matrices obtained from combinations of snapshots for different dynamic viscosity parameter $\mu$ using the POD approach. 

The anchor points are annotated and their respective clusters are color-coded. We then obtain the global projection matrix $[\Phi^*]$ similarly, using the POD approach. Fig.~\ref{fig: errorFunc_Global_NS_r3} shows the graph of the error function $r \mapsto \epsilon(r)$ for the anchor points in $\mathcal{A}$ and the global projection matrix $[\Phi^*]$. With the same error threshold, $r=3$ is selected to build the anchor points and the global projection matrix $[\Phi^*]$, leading to stochastic modeling in $\mathbb{S}_{85808, {(3+13)}} \subset \text{St}_{85808, {(3+13)}}$.

\begin{figure}[tbp]
    \centering
    \begin{subfigure}{0.45\textwidth}
    \includegraphics[width=\textwidth]{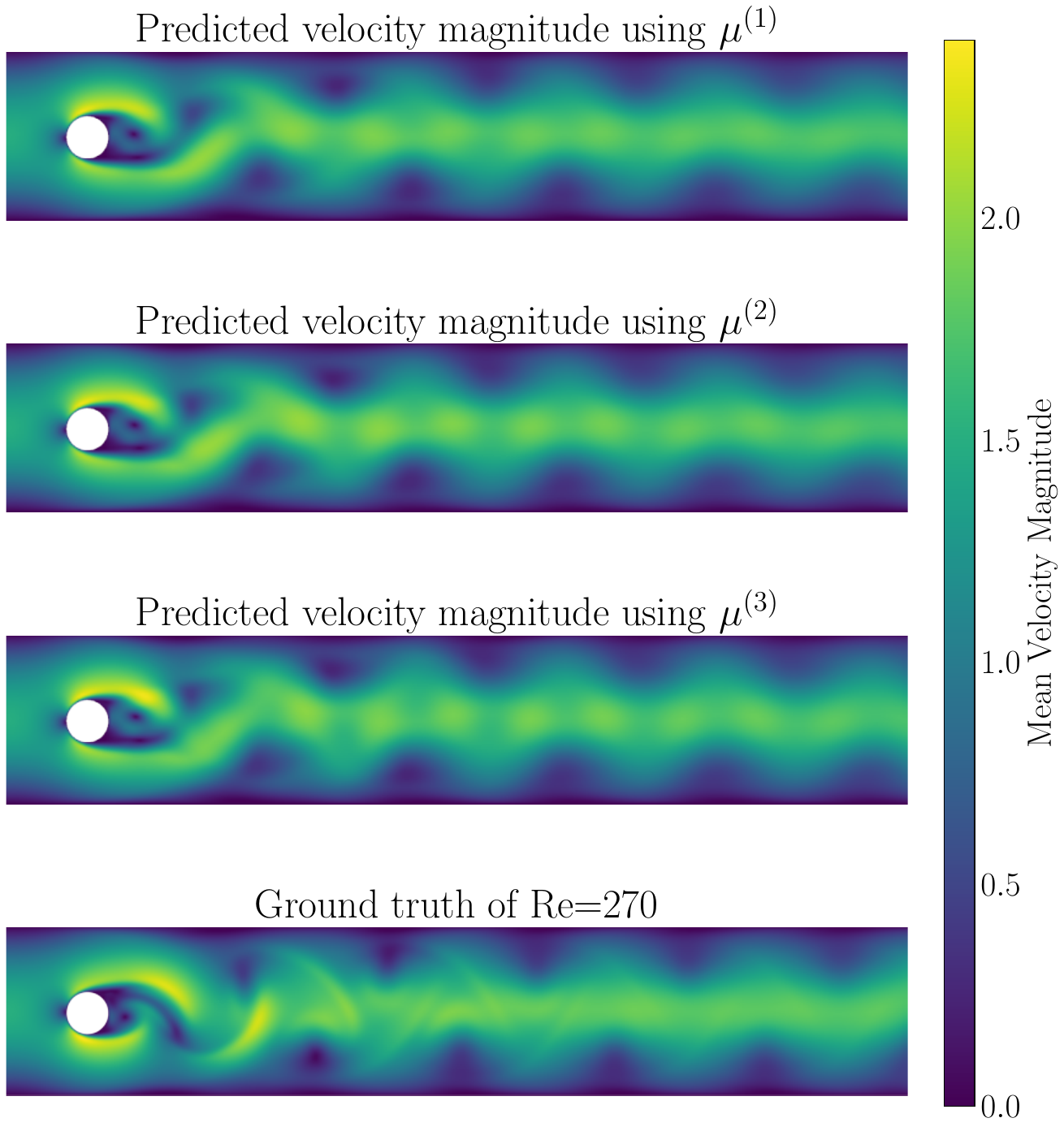}
    \caption{Velocity magnitude}
    \label{fig: soln_comb_selected_operators_velocity_mag}
    \end{subfigure} \hspace{1em}
    \begin{subfigure}{0.45\textwidth}
    \includegraphics[width=\textwidth]{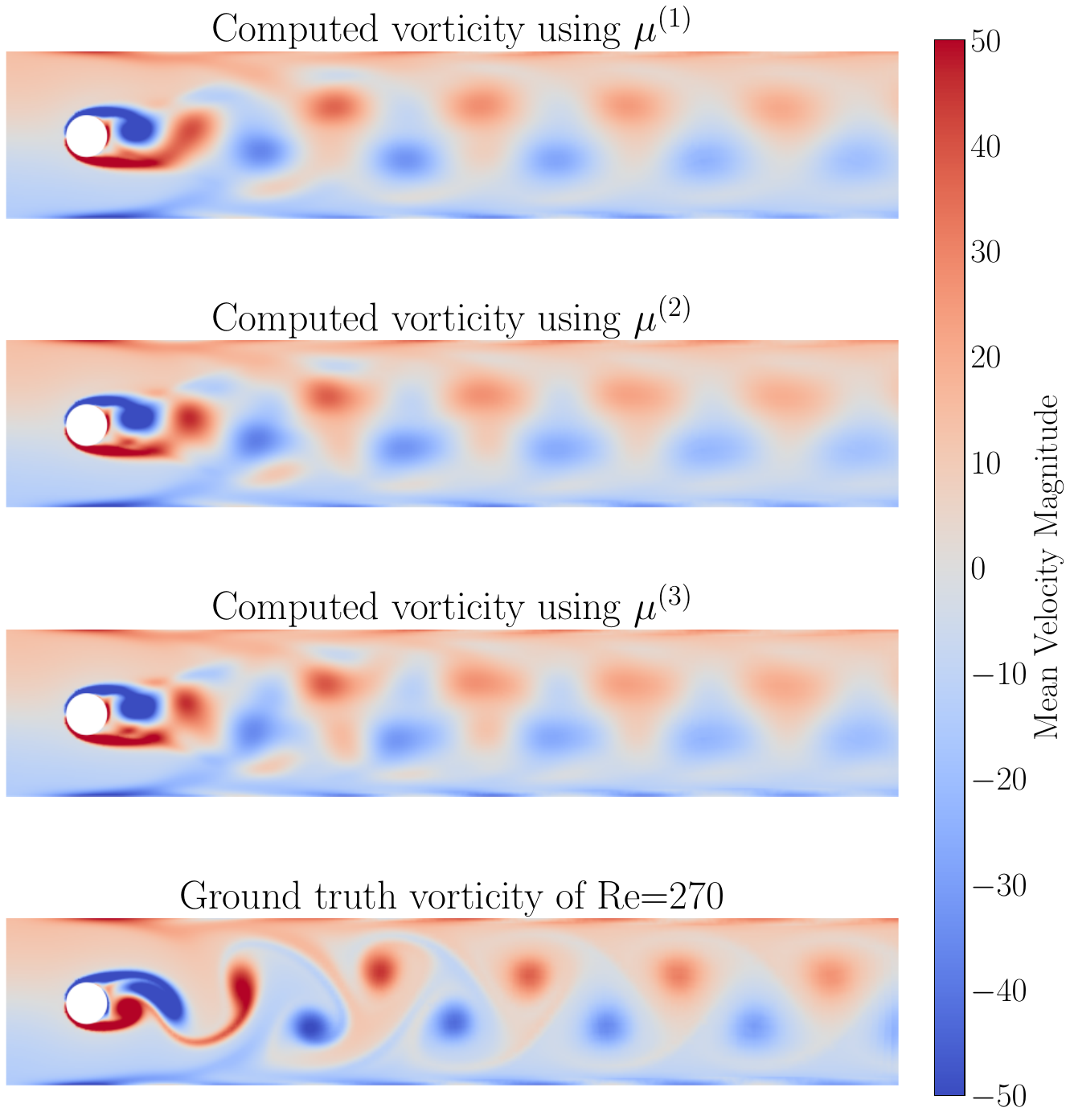}
    \caption{Vorticity}
    \label{fig: soln_comb_selected_operators_vorticity}
    \end{subfigure}
    \caption{Velocity magnitude (left) and vorticity (right) of the reconstructed velocity states of the two-dimensional transient flow past circular cylinder problem \eqref{eqn: NS2D} at time $t=5$s for anchor points $\mathcal{A}$ corresponding to $\mu^{(1)}$, $\mu^{(2)}$ and $\mu^{(3)}$ using their respective inferred operators. Note that the vorticity colormap is truncated at $\pm50\text{s}^{-1}$ to enhance visualization of vortex shedding patterns.}
\end{figure}

Fig.~\ref{fig: soln_comb_selected_operators_velocity_mag} and Fig.~\ref{fig: soln_comb_selected_operators_vorticity} present the computed velocity magnitude and vorticity from the reconstructed velocity states. 

\begin{figure}[tbp]
    \centering
\end{figure}

The velocity states are reconstructed with the operators obtained from different combinations of the viscosity parameter $\mu$. We observe that both the computed velocity magnitude and vorticity of the anchor points do not deviate significantly from each other, maintaining consistency across the anchor points. A notable observation from this study is that an increase in data quantity does not necessarily correlate with improved accuracy in computed vorticity patterns. As evident in Fig.~\ref{fig: soln_comb_selected_operators_vorticity}, the vorticity computed using $\mu^{(1)}$, which utilizes a smaller dataset, demonstrates a closer resemblance to the ground truth vorticity pattern compared to that computed using $\mu^{(2)}$, which employs a larger dataset. This finding underscores the critical role of data composition in determining the model's predictive performance. It suggests that the variability introduced by the data mixture significantly influences the model's predictive capabilities, emphasizing the importance of optimizing data selection and integration strategies to enhance overall model performance.

\begin{figure}[tbp]
    \centering
    \includegraphics[width=\textwidth]{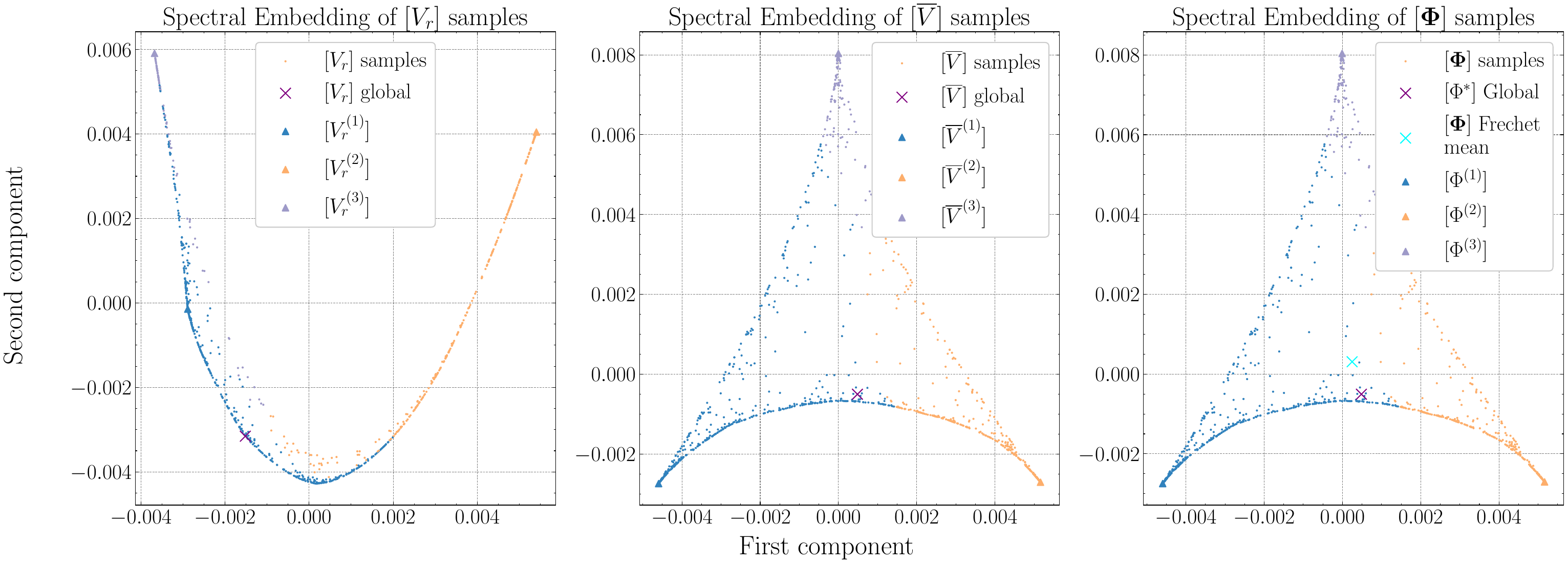}
    \caption{Visualization of the dataset, given by the three anchor points corresponding to $\mu^{(1)}$, $\mu^{(2)}$, and $\mu^{(3)}$, and the 1000 generated samples of the stochastic projection matrix $[\mathbf{\Phi}]$. Spectral embedding is used to project data in $\text{St}(85808, 3+13)$ to $\mathbb{R}^{2}$. Left: Spectral embedding visualization of $[\bfV_r]$ Middle: Spectral embedding visualization of $[\overline{\bfV}]$. Right: Spectral embedding visualization of the stochastic projection matrix $[\mathbf{\Phi}]$.}
    \label{fig: SE_StiefelSamples_KMeans_NS_q13_r3}
\end{figure}

Regarding the randomization on the tangent space, the concentration parameters $\boldsymbol{\alpha}$ of the Dirichlet distribution are obtained as $\boldsymbol{\alpha} = (0.4967, 0.3591, 0.1442)^\top$. Samples of the stochastic projection matrix $[\mathbf{\Phi}]$ are shown in Fig.~\ref{fig: SE_StiefelSamples_KMeans_NS_q13_r3}. 
As expected, the 1,000 generated samples are located within the convex hull defined by the three anchor points. The Fr\'echet mean of the samples is seen to be reasonably close to the global projection matrix $[\Phi^*]$ in the reduced space, as shown in the rightmost panel in Fig.~\ref{fig: SE_StiefelSamples_KMeans_NS_q13_r3}. We recall that such visual representations are obtained through a nonlinear reduction technique and are only used to provide qualitative insight. 

\begin{figure}[tbp]
    \centering
    \includegraphics[width=.95\textwidth]{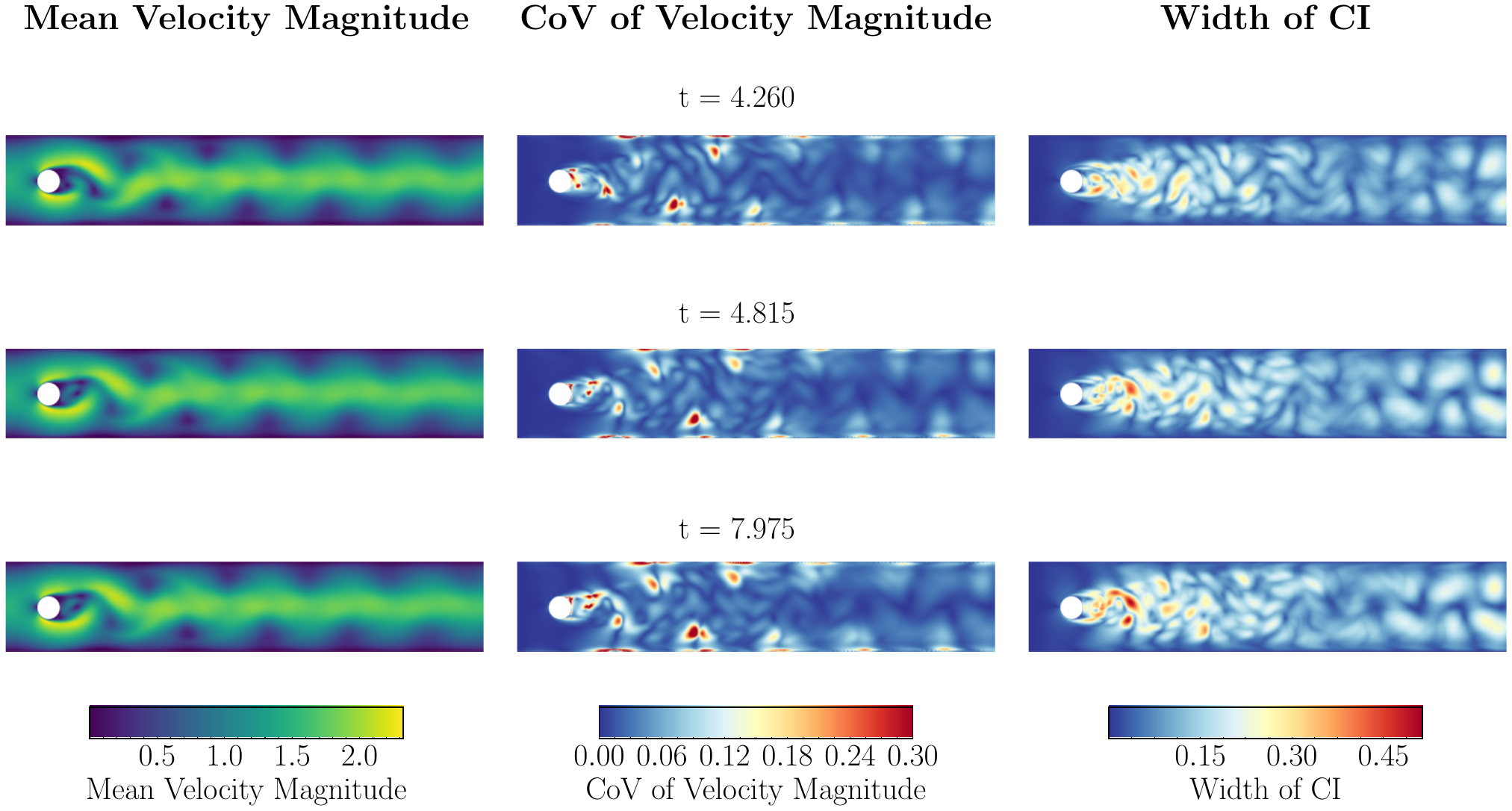}
    \caption{Mean velocity magnitude and uncertainty measures at various time steps. Left: The mean velocity magnitude obtained using the stochastic reduced-order model. Middle: Coefficient of Variation (CoV) obtained using the stochastic reduced-order model. To facilitate visual interpretation, CoV values are truncated at the 99th percentile of the distribution, with higher values mapped to this upper threshold. Right: The width of the 95\% confidence interval of the velocity magnitude obtained using the stochastic reduced-order model.} 
    \label{fig: magnitude_uncertainty_plot_pyvista}
\end{figure}
\begin{figure}[tbp]
    \centering
    \includegraphics[width=.95\textwidth]{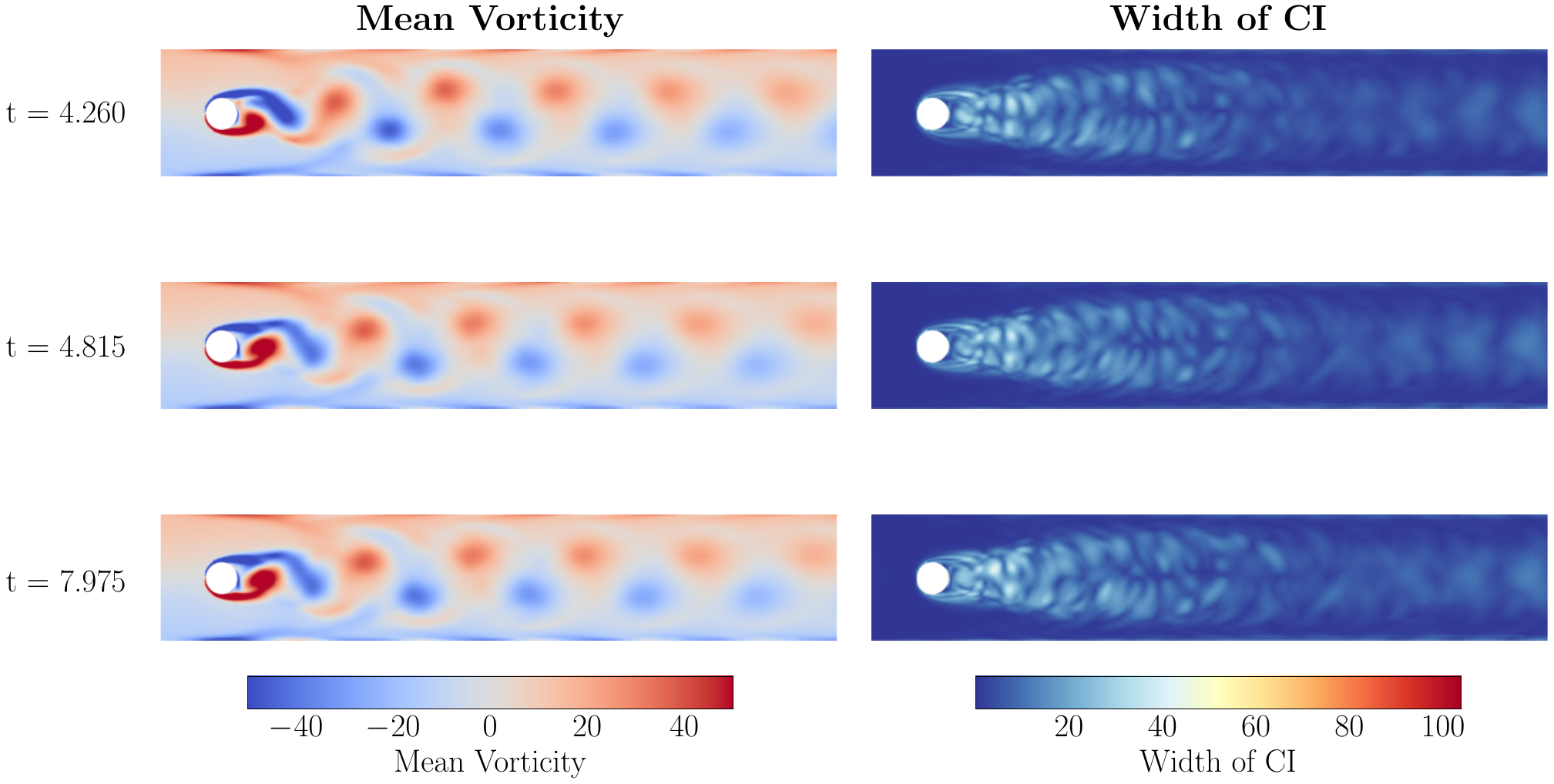}
    \caption{Temporal evolution of mean vorticity and associated width of the 95\% confidence interval. The mean vorticity colormap is truncated at $\pm50\text{s}^{-1}$ to enhance visualization of vortex shedding patterns. Left: The mean vorticity obtained using the stochastic reduced-order model. Right: The width of the 95\% confidence interval of the vorticity obtained using the stochastic reduced-order model.}
    \label{fig: vorticity_uncertainty_plot_pyvista}
\end{figure}

\begin{figure}[tbp]
    \centering
    \begin{subfigure}{0.45\textwidth}
    \includegraphics[width=\textwidth]{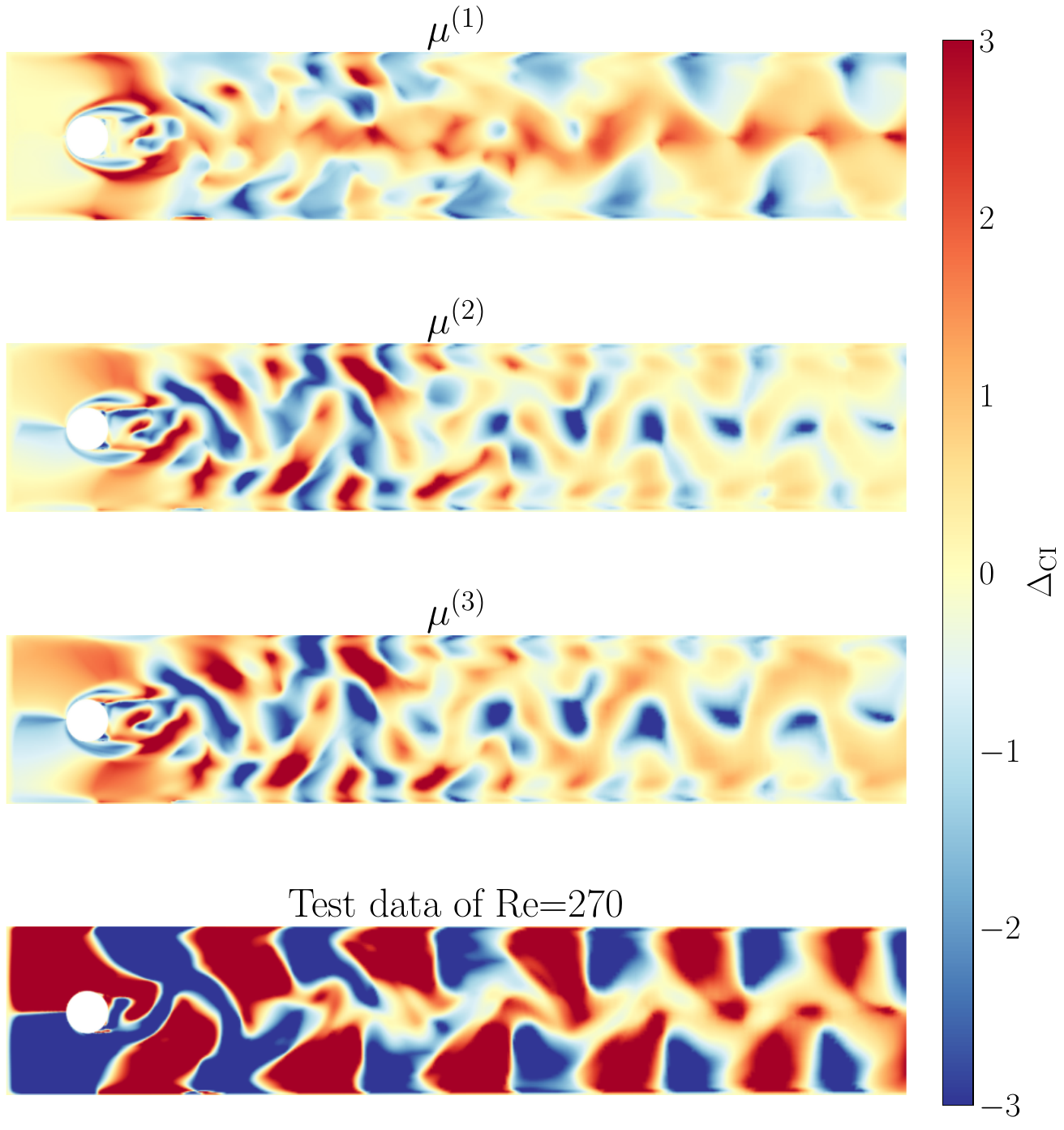}
    \caption{Velocity magnitude}
    \label{fig: u_s_rec_full_operator_compare_CI}
    \end{subfigure} \hspace{1em}
    \begin{subfigure}{0.45\textwidth}
    \includegraphics[width=\textwidth]{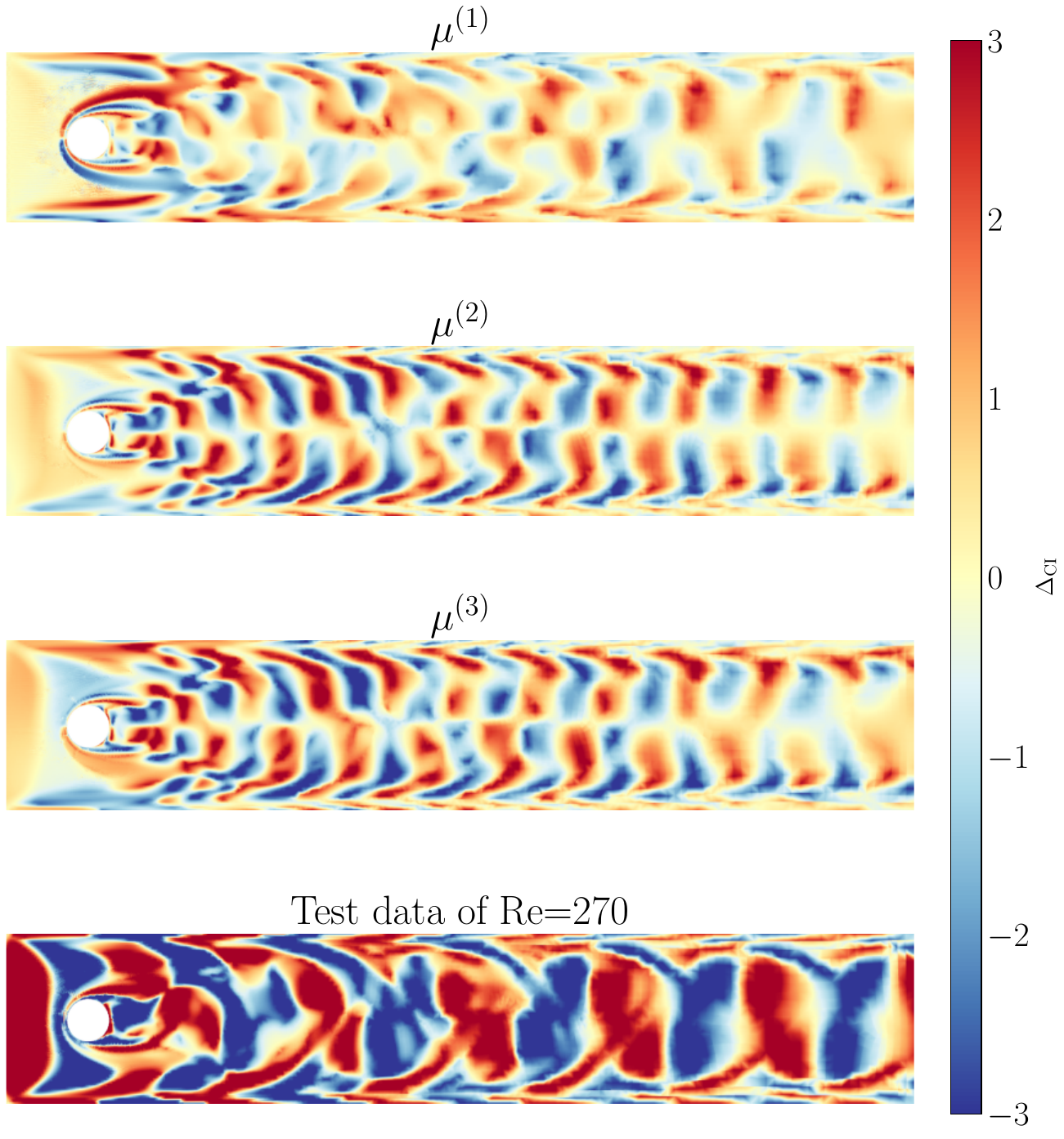}
    \caption{Vorticity}
    \label{fig: vorticity_s_rec_full_operator_compare_CI}
    \end{subfigure}
    \caption{Spatial distribution of velocity magnitude (left) and vorticity (right) deviations from the mean velocity magnitude and vorticity respectively for anchor points $\mathcal{A}$ corresponding to $\mu^{(1)}$, $\mu^{(2)}$ and $\mu^{(3)}$. The deviation $\Delta_{\text{CI}}$ is expressed in terms of one half the confidence interval width (see Eq.~\ref{eqn: deltaCI}), where $\Delta_{\text{CI}} = \pm1$ coincides with the boundaries of the 95\% confidence interval. Values of $\left|\Delta_{\text{CI}} \right| < 1$ indicate model responses within the 95\% confidence interval, while $\left|\Delta_{\text{CI}} \right| > 1$ represent responses outside this interval. Dark red regions $\Delta_{\text{CI}} > 1$) indicate areas where the anchor point model predicts higher vorticity values than the upper bound of the stochastic reduced-order model's 95\% confidence interval. Conversely, dark blue regions ($\Delta_{\text{CI}} < -1$) signify areas where the anchor point model predicts lower vorticity values than the lower bound of the stochastic reduced-order model's 95\% confidence interval.}
\end{figure}

Finally, model-form uncertainties are propagated through Monte Carlo simulations. Fig.~\ref{fig: magnitude_uncertainty_plot_pyvista} and Fig.~\ref{fig: vorticity_uncertainty_plot_pyvista} illustrate the width of the 95\% confidence interval for the velocity magnitude and vorticity fields, respectively, obtained using the proposed probabilistic formulation. To enable efficient visualization of the 95\% confidence interval of individual samples obtained using our stochastic reduced-order model, we calculate the width of the 95\% confidence interval for each quantity of interest (QoI) using the percentile method:
\begin{equation}
    \begin{aligned}
        \text{Width}_{\text{CI}} &= P_{97.5} - P_{2.5}\,,
    \label{eqn: CIwidth}
    \end{aligned}
\end{equation}
where $P_{97.5}$ is the 97.5th percentile of the QoI (velocity magnitude or vorticity) samples and $P_{2.5}$ is the 2.5th percentile of the QoI samples computed across the generated samples. To evaluate the performance of our stochastic reduced-order model, we compare its 95\% confidence intervals with the responses from anchor point reduced-order models for a test case at $\text{Re}=270$ (Fig.\ref{fig: u_s_rec_full_operator_compare_CI}). We quantify the deviation of anchor point responses relative to the confidence interval using the metric:
\begin{equation}
    \Delta_{\text{CI}} = \frac{\text{QoI} - \overline{\text{QoI}}}{0.5\text{Width}_{\text{CI}}}\,,
    \label{eqn: deltaCI}
\end{equation}
where $\overline{\text{QoI}}$ denotes the mean of the quantity of interest. This normalized deviation allows us to assess how well the stochastic model encompasses the deterministic predictions from different anchor points.
Our analysis reveals that for the $\mu^{(1)}$ anchor point, the stochastic model demonstrates good agreement, with the majority of the domain exhibiting $\left|\Delta_{\text{CI}}\right| < 1$. This indicates that the $\mu^{(1)}$ anchor point response largely falls within the 95\% confidence interval of our stochastic model. In contrast, the $\mu^{(2)}$ and $\mu^{(3)}$ anchor point models show more extensive regions where $\left|\Delta_{\text{CI}}\right| > 1$, particularly in the vicinity of the circular obstacle and within the wake region. This suggests that our stochastic reduced-order model generates samples that deviate significantly from these anchor point responses in areas where vortex shedding patterns occur (see Fig.~\ref{fig: soln_comb_selected_operators_velocity_mag} and Fig.~\ref{fig: soln_comb_selected_operators_vorticity}). Notably, we observe that some stochastic samples extend beyond the domain defined by the nominal responses, especially for $\mu^{(2)}$ and $\mu^{(3)}$. This highlights the model's capacity to explore a wider range of potential flow behaviors than captured by individual deterministic simulations.
When comparing to the ground truth data for $\text{Re}=270$, we note significant deviations from all nominal responses. Areas associated with vortex shedding patterns exhibit $\left|\Delta_{\text{CI}}\right| > 1$, highlighting the challenges in accurately modeling the peaks and troughs of the flow field. These observations demonstrate the stochastic reduced-order model's ability to generate samples that not only align well with the anchor point model responses but also able to capture flow variations beyond those predicted by the deterministic anchor point models.

The analysis of Fig.~\ref{fig: magnitude_uncertainty_plot_pyvista} indicates variations in velocity magnitudes, particularly near the obstacle and in the wake region. The Coefficient of Variation (CoV) is displayed with values truncated at the 99th percentile of the distribution, with higher values mapped to this upper threshold of 0.30. This truncation helps visualize the overall uncertainty pattern without being skewed by extreme values. The highest visible CoV values occur near the obstacle and in certain areas of the wake. The width of the 95\% confidence interval follows a similar pattern, with maximum visible values around 0.6 at regions proximal to the obstacle and in parts of the wake. A subtle temporal trend is observed wherein these uncertainty measures exhibit slight increases as the simulation progresses from $t = 4.260$s to $t = 7.975$s, most notably in the wake region. 

Fig.~\ref{fig: vorticity_uncertainty_plot_pyvista} illustrates the temporal evolution of mean vorticity and the associated width of the 95\% confidence interval for vorticity. The mean vorticity colormap is truncated at $\pm50\text{s}^{-1}$ to enhance visualization of the vortex shedding patterns. The width of the 95\% confidence interval for vorticity reveals significantly higher uncertainties compared to the velocity magnitude, particularly in the wake region behind the obstacle. These uncertainty values reach up to $100\text{s}^{-1}$, indicating substantial variability in vorticity predictions. The regions of elevated uncertainty extend downstream, with two distinct areas of heightened variability: immediately behind the obstacle, where initial vortex formation occurs, and further downstream, where complex vortex interactions are likely to take place. As the simulation progresses from $t = 4.260$s to $t = 7.975$s, a subtle yet noticeable expansion of the high-uncertainty region is observed in the wake. This expansion suggests a gradual decrease in the model's predictive confidence for vorticity as the simulation approaches later time steps. Despite this temporal evolution, the overall spatial distribution of uncertainty maintains a relatively consistent pattern throughout the simulation, with the highest uncertainties consistently localized in the wake region behind the obstacle.

\section{Conclusions}
\label{sec:conclusions}

A probabilistic framework enabling the quantification of uncertainties in operator inference has been proposed. The approach combines stochastic reduced-order modeling where the projection basis is appropriately randomized on a matrix manifold, and a methodology to promote consistency at the inference stage. The efficiency of the approach was assessed through numerical experiments on Burgers' and two-dimensional Navier-Stokes equations. Specifically, the stochastic method was deployed to investigate the robustness of the operator inference framework with respect to the training strategy. It was shown that the proposed technique allow for the fluctuations introduced while selecting the training parameters---which can be as large as ten percents---can be captured. 

The proposed stochastic reduced-order modeling approach, while effective, presents certain limitations that deserve consideration. The computational complexity associated with the Riemannian operators for projection and retraction on the Stiefel manifold can prove substantial, particularly for systems with high-dimensional state spaces or when a large number of basis vectors are retained. Additionally, although a systematic approach for anchor points selection has been developed, the choice of these points may not be optimal, which may impact the robustness of our UQ results.

Perspectives for future work include (i) extensions to inference and reduced-order modeling involving nonlinear closure terms and alternative linear subspace strategies, (ii) methodological developments ensuring consistency for arbitrarily chosen reduced-order operators, (iii) goal-oriented optimal design to identify anchor points, and (iv) robustness analyses with respect to other training parameters. 

\section*{Acknowledgements}
The work of J.\ Y.\ Y.\ and J.\ G.\ was supported by the National Science Foundation under awards DGE-2022040 and CMMI-1942928, and by the Army Research Office under grant W911NF-23-1-0125.

\bibliographystyle{abbrv}
\bibliography{Bib}

\end{document}